\documentclass{article}

\PassOptionsToPackage{numbers, compress}{natbib}
\usepackage[final]{neurips_2024}

\usepackage[utf8]{inputenc} % allow utf-8 input
\usepackage[T1]{fontenc}    % use 8-bit T1 fonts
\usepackage{hyperref}       % hyperlinks
\usepackage{url}            % simple URL typesetting
\usepackage{booktabs}       % professional-quality tables
\usepackage{amsfonts}       % blackboard math symbols
\usepackage{nicefrac}       % compact symbols for 1/2, etc.
\usepackage{microtype}      % microtypography
\usepackage[table]{xcolor}         % colors
\usepackage{subfig}
\usepackage{pifont}
\usepackage{enumitem}
\usepackage{listings}
\usepackage{algorithm}
\usepackage{algpseudocode}
\usepackage{wrapfig}
\usepackage{tcolorbox}

\usepackage{graphicx}
\usepackage{multirow}
\usepackage{amsmath}
\usepackage{amssymb}
\usepackage{mathtools}
\usepackage{amsthm}
\usepackage{dsfont}
\usepackage{array}

\usepackage[toc,page,header]{appendix}
\usepackage{minitoc}

\title{ReLU's Revival: On the Entropic Overload in Normalization-Free Large Language Models}

\author{%
  Nandan Kumar Jha \\
  New York University\\
  \texttt{nj2049@nyu.edu} \\
   \And
  Brandon Reagen \\
  New York University\\
  \texttt{bjr5@nyu.edu} \\
}

\begin{document}

\doparttoc % Tell to minitoc to generate a toc for the parts
\faketableofcontents % Run a fake tableofcontents command for the partocs

\part{} % Start the document part
%\parttoc % Use this only when you want to have ToC for main paper, too

\maketitle

\begin{abstract}
LayerNorm is a critical component in modern large language models (LLMs) for stabilizing training and ensuring smooth optimization. However, it introduces significant challenges in mechanistic interpretability, outlier feature suppression, faithful signal propagation, and computational and communication complexity of private inference. This work explores desirable activation functions in normalization-free decoder-only LLMs. Contrary to the conventional preference for the GELU in transformer-based models, our empirical findings demonstrate an {\em opposite trend}---ReLU significantly outperforms GELU in LayerNorm-free models, leading to an {\bf 8.2\%} perplexity improvement.  We discover a key issue with GELU, where early layers experience entropic overload, leading to the under-utilization of the representational capacity of attention heads. This highlights that smoother activations like GELU are {\em ill-suited} for LayerNorm-free architectures, whereas ReLU's geometrical properties---specialization in input space and intra-class selectivity---lead to improved learning dynamics and better information retention in the absence of LayerNorm. This study offers key insights for optimizing transformer architectures where LayerNorm introduces significant challenges. The code and implementation are available at \href{https://github.com/Nandan91/relu-revival-normfree}{relu-revival-normfree}.
\end{abstract}

\section{Introduction}

{\bf Motivation and challenges.}
LayerNorm \cite{ba2016layer} has been a key architectural component contributing to the success of large language models (LLMs) by stabilizing training through normalizing inputs across features within a layer. Additionally, it plays a crucial role in enhancing the models' non-linear representational capabilities \citep{wu2024role,ni2024on,zhao2023tuning,joudaki2023impact}. Despite its benefits, LayerNorm introduces several practical challenges that become pronounced in specific settings:
\begin{enumerate} [noitemsep,nolistsep,leftmargin=0.5cm] \vspace{-0.5em}
\item {\em Private Inference (PI):} PI protocols enable inference on encrypted data without exposing inputs, ensuring data privacy while protecting model weights   \citep{zhang2024secure,lu2023bumblebee,zimerman2023converting,pang2023bolt,gupta2023sigma,hou2023ciphergpt,jha2023deepreshape,ghodsi2021circa,jha2021deepreduce}. 
Hybrid PI protocols encounter difficulties with the inverse square root computation inherent in LayerNorm, making it the second most costly operation after GELU, contributing to 22\% of total latency and communication overhead \citep{hou2023ciphergpt}. Also, Homomorphic Encryption (HE)-only PI requires polynomial approximations of LayerNorm, which are challenging due to the wide variance range \citep{zimerman2023converting}.
\item {\em Mechanistic Interpretability:} LayerNorm increases the complexity of the residual stream, making it harder to analyze and understand the internal workings of transformer models \citep{nanda2023attribution}, limiting the applicability of LLMs for applications requiring transparency and explainability. 
\item {\em Low-Precision Training:} The trainable parameters in LayerNorm are associated with the amplification of outlier features which poses challenges in LLM quantization, as it exacerbates numerical instability and degrades performance in low-precision training regimes \citep{he2024understanding,bondarenko2024quantizable,wei2022outlier,kovaleva2021bert}.
\item {\em Signal Propagation:} LayerNorms shown to negatively impact the faithful signal propagation \cite{he2023deep}.
\end{enumerate}

These challenges highlight the need for LayerNorm-free architectures that preserve transformer benefits while avoiding its drawbacks. However, this shift introduces new considerations, especially in selecting activation functions for the feed-forward networks (FFNs). Prior work \citep{he2023deep,noci2024shaped,he2024simplifying} has explored various architectural heuristics for designing normalization-free LLMs. However, the impact of removing normalization layers on the choice of FFN activation functions remains underexplored.

\begin{figure} [t]
\centering
\subfloat[SM + LN + G \label{subfig:BaselineGELU}]{\includegraphics[width=.25\textwidth]{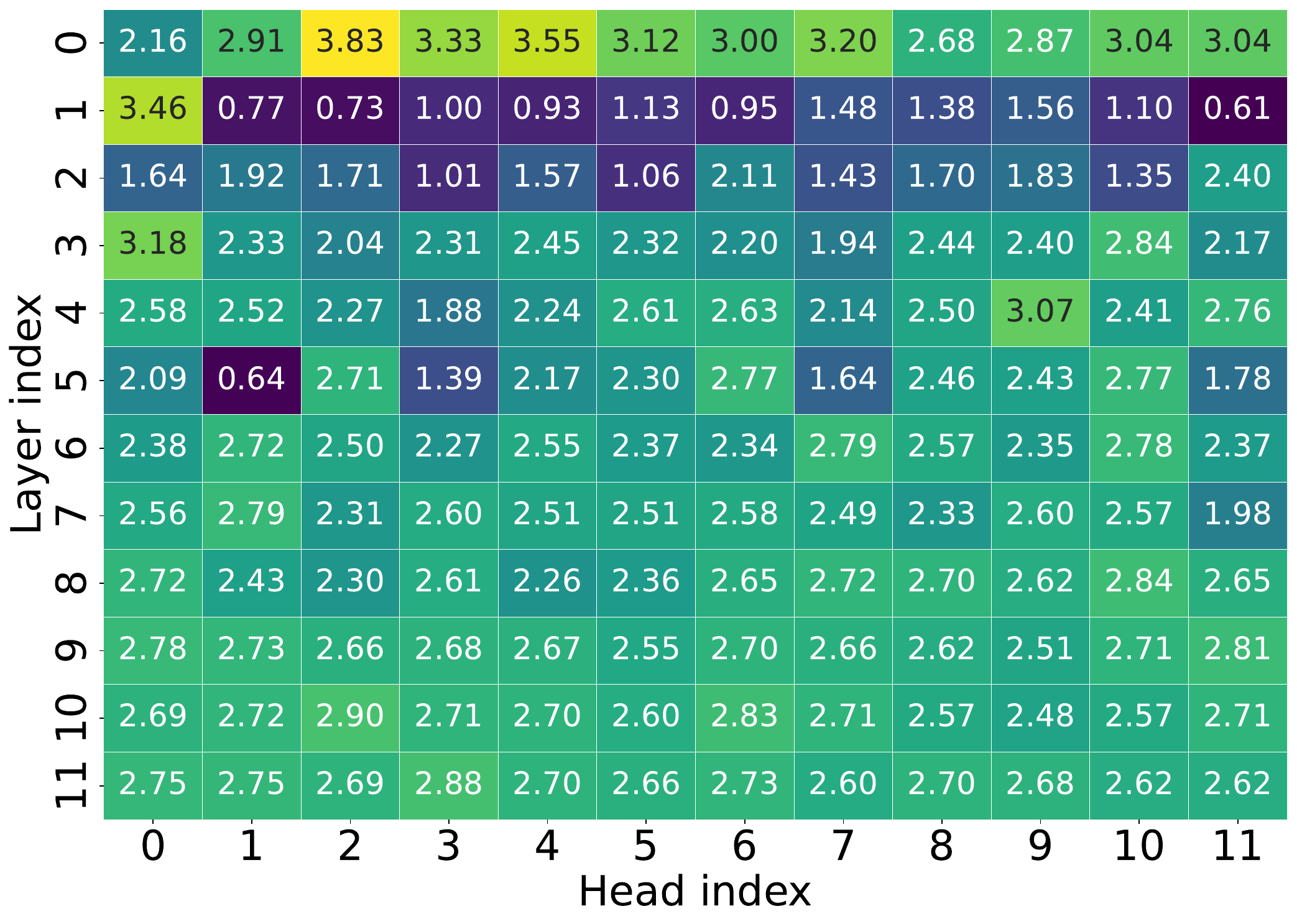}}
\subfloat[SM + LN + R \label{subfig:BaselineReLU}]{\includegraphics[width=.25\textwidth]{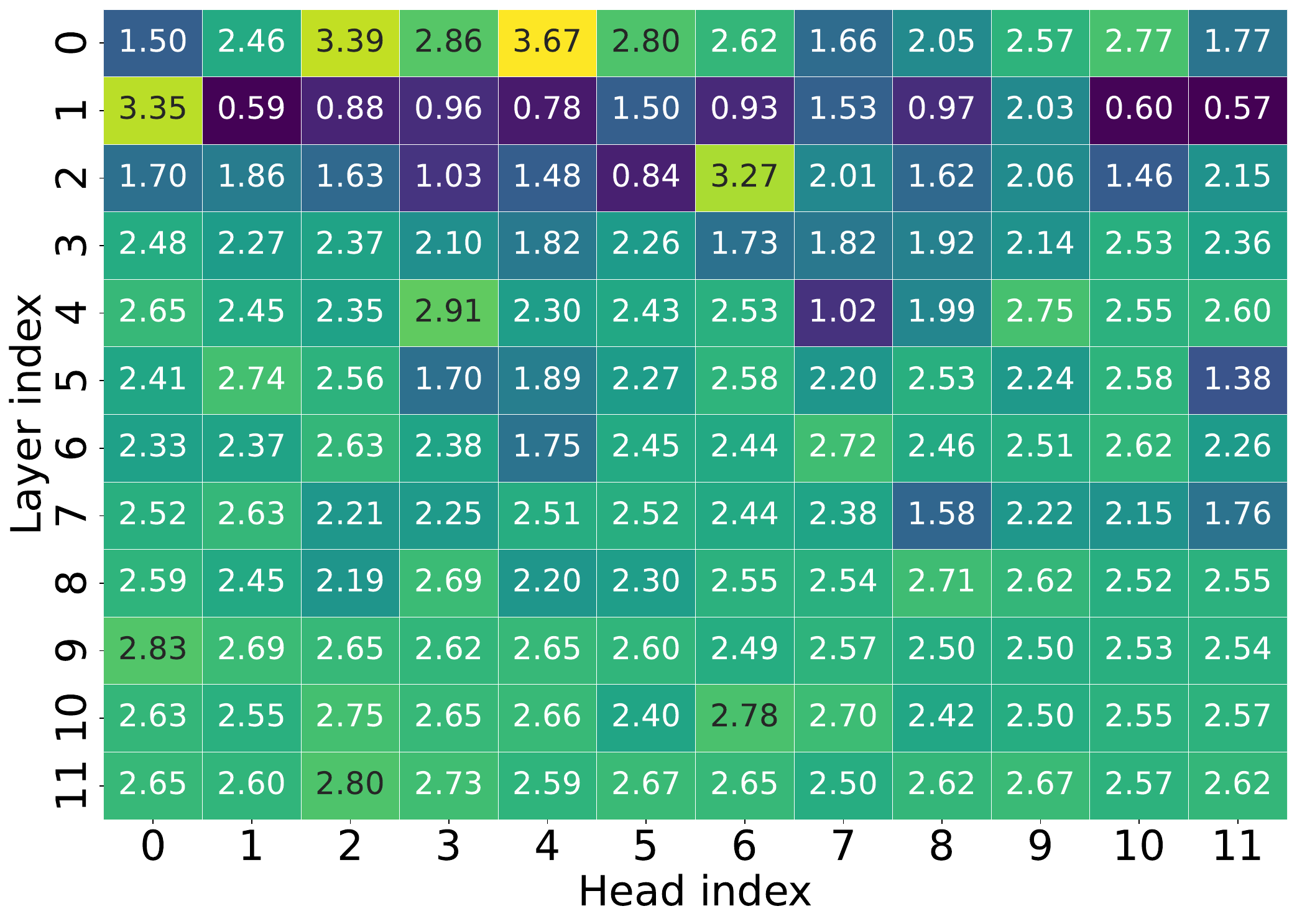}} 
\subfloat[SM + G]{\includegraphics[width=.25\textwidth]{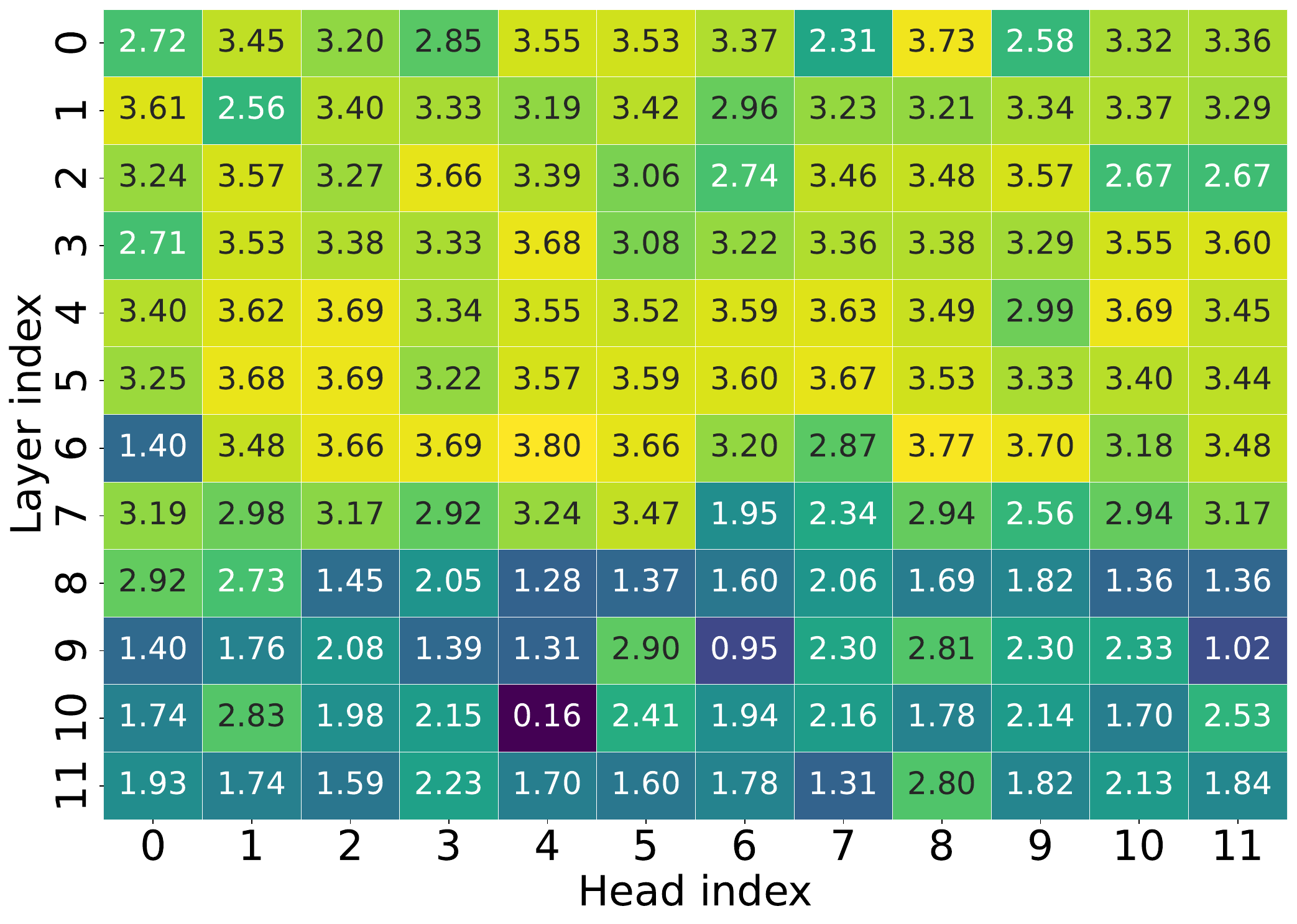}} 
\subfloat[SM + R]{\includegraphics[width=.25\textwidth]{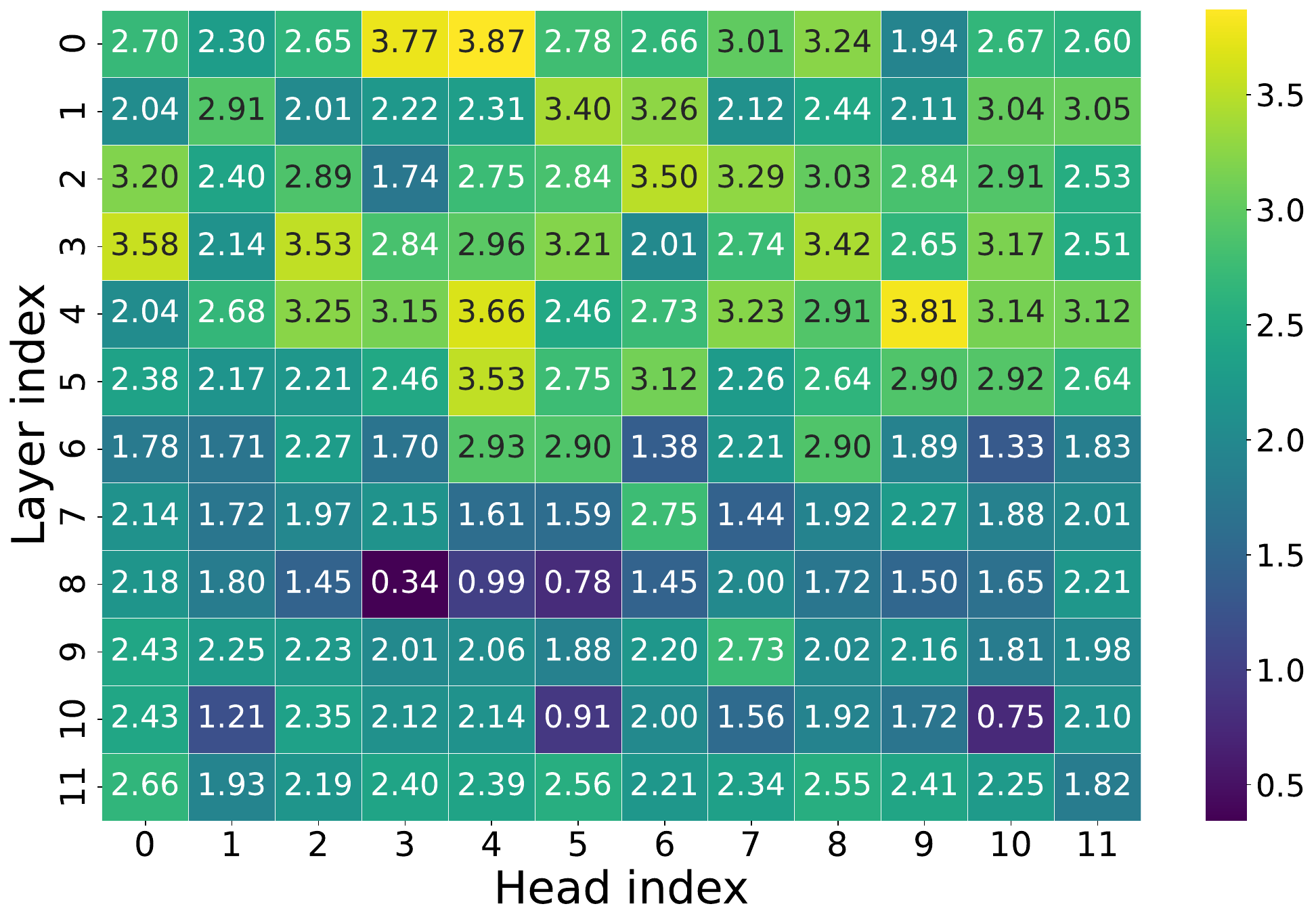}} 
\vspace{-0.5em}
\caption{Entropy heatmaps of attention for baseline (a, b) and normalization-free (c, d) GPT-2 models with GELU and ReLU in the FFN. In the absence of LayerNorm, GELU in the FFN leads to significantly higher entropic overload (highlighted in yellow, c) compared to ReLU.}
\label{fig:AttnEntHeatMaps}
\end{figure}

{\bf Research insights and its implications.}
In this work, we go beyond prior approaches by conducting an in-depth investigation into the design choices for activation functions in normalization-free LLMs, offering new insights into how these choices impact learning dynamics, internal representations, and overall model performance. Our study reveals several key findings:
\begin{itemize} [noitemsep,nolistsep,leftmargin=0.5cm] \vspace{-0.5em}
\item {\em ReLU Outperforms GELU in LayerNorm-Free Models:} Contrary to conventional practices, we show that models using ReLU in the FFN significantly outperform, with an 8.2\% improvement in perplexity, those using GELU in the absence of LayerNorm (see Figure \ref{fig:LossCurveNonlinConfig} and Table \ref{tab:NonlinConfigEvalPPL}). 
\item {\em Learning Dynamics with Learnable Negative Slopes:} To explore further, we experimented with a learnable negative slope in the leaky ReLU activation function using two configurations: (1) Layer-wise configuration: Each layer has its independent learnable slope. Initially, early layers learn a positive slope while deeper layers learn a negative slope. However, gradually all layers converge to a near-zero slope (Figure \ref{fig:LearnableNegSlope}a). (2) Global Configuration: A single learnable slope is shared across all layers. The slope initially shifts to positive before converging to near zero (see Figure \ref{fig:LearnableNegSlope}b). These results highlight LayerNorm-free models' inherent preference for ReLU-like activations with zero negative slopes.
\item {\em Entropic Overload with GELU Activation:} To delve deeper, we analyze the head-wise entropy values and find that early layers in normalization-free models with GELU activation experience entropic overload, a significant proportion of attention heads reach near-maximum entropy levels, indicating the under-utilization of the representational capacity of attention heads. 
\end{itemize}

{\bf Contributions.} Our key contributions are follows: \vspace{-0.5em} 
\begin{enumerate} [noitemsep,nolistsep,leftmargin=0.5cm] 
\item We conduct an in-depth analysis of activation functions in normalization-free, decoder-only models by studying their learning dynamics when trained from scratch. 

\item We explore the effect of different activation functions in baseline and normalization-free models on the attention score distribution through the lens of Shannon's entropy, offering valuable insights for advancing the architectural design of LayerNorm-free models.

\item We conducted experiments across various context sizes (128 and 256) on GPT-2 and Pythia \cite{biderman2023pythia} model with 2.1B training tokens from the CodeParrot \cite{codeParrot}. 
\end{enumerate}

\section{Preliminaries}

{\bf Notations.}
We denote the number of layers as $L$, number of heads as $H$, model dimensionality as $d$, head dimension as $d_k$ (where \(d_k = \frac{d}{H}\)), and context length as $T$.  Table \ref{tab:ArchConfigGPT2} illustrates the abbreviations for architectural configurations with simplified nonlinearities in a transformer-based LLM.

\subsection{Overview of Transformer-based Decoder-only Architectures}

%{\bf An overview of transformer-based decoder-only architecture.}
A transformer-based LLM is constructed by sequentially stacking \(L\) transformer blocks, where each block is composed of two sub-blocks: an attention mechanism and a feed-forward network (FFN), both having their own residual connections and normalization layers, positioned in the Pre-LN order to improves training stability \citep{xiong2020layer}. Formally, transformer blocks take an input sequence \(\mathbf{X}_{\text{in}} \in \mathbb{R}^{T \times d}\), consisting of \(T\) tokens of dimension \(d\), and transform it into \(\mathbf{X}_{\text{out}}\) as follows:

\vspace{-1.5em}

\begin{equation} \label{eqn:ffn_mha}
\mathbf{X}_{\text{out}} = \hat{\mathbf{X}}_{\text{SA}} + \text{FFN}_{\text{GELU}}(\text{LayerNorm}_2(\hat{\mathbf{X}}_{\text{SA}})), \; \text{where} \; \hat{\mathbf{X}}_{\text{SA}} = \mathbf{X}_{\text{in}} + \text{MHA}(\text{LayerNorm}_1(\mathbf{X}_{\text{in}})).
\end{equation}

The Multi-Head Attention (MHA) sub-block enables input contextualization by sharing information between individual tokens. MHA employs the self-attention mechanism to compute the similarity score of each token with respect to all other tokens in the sequence. In particular, self-attention mechanism transform the input sequence \(\mathbf{X}\) into  \(\mathbf{Attn}(\mathbf{X})\) as follows:

\vspace{-1.5em}

\begin{equation} \label{eqn:attn_softmax}
\mathbf{Attn}(\mathbf{X}) = \Big(\text{Softmax}\Big(\frac{1}{\sqrt{d_k}} (\mathbf{X} \mathbf{W}^Q) (\mathbf{X}{\mathbf{W}^K})^\top + \mathbf{M}\Big)\Big)\mathbf{X}\mathbf{W}^V.
\end{equation}

Here, each token generates query($Q$), key($K$), and value($V$) vectors through the linear transformations \(\mathbf{W}^Q, \mathbf{W}^K, \; \text{and} \; \mathbf{W}^V \in \mathbb{R}^{d \times d_h}\), respectively. Then, similarity scores are computed by taking the dot product of the $Q$ and $K$ vectors, scaled by the inverse square root of the $K$ dimension, and passed through a softmax function to obtain the attention weights. These weights are then used to compute a weighted sum of the $V$ vectors, producing the output for each token. For auto-regressive models (e.g., GPT), mask \(\mathbf{M} \in \mathbb{R}^{T\times T}\),  which has values in \(\{0, -\infty\}\) with \(\mathbf{M}_{i,j} = 0 \,\text{iff} \, {i \geq j}\), is deployed to prevent the tokens from obtaining information from future tokens.

The MHA sub-block employs a self-attention mechanism across all the heads, each with its own sets of $Q$, $K$, and $V$. This allows the attention heads to focus on different parts of the input sequence, capturing various aspects of the input data simultaneously. The outputs from all heads are concatenated and linearly transformed ($\mathbf{W}^O\in\mathbb{R}^{d\times d}$) to produce the final MHA output as follows: 

\vspace{-1.3em}

\begin{equation} \label{eqn:mha_concat}
\text{MHA}(\mathbf{X}) = \text{Concat}\big(\text{Attn}_1(\mathbf{X}), \; \text{Attn}_2(\mathbf{X}), \; \text{Attn}_3(\mathbf{X}), \dots, \text{Attn}_H(\mathbf{X})\big) \mathbf{W}^O.
\end{equation}

Following the MHA sub-block, the FFN sub-block transforms each token independently. The FFN sub-blocks have a single hidden layer whose dimension is a multiple of \(d\) (e.g., \(4d\) in GPT \citep{radford2019language}  models). Specifically, the FFN sub-block first applies a linear transformation to the input \(\mathbf{X}\) using \(\mathbf{W}^{\text{ffn}}_{\text{in}} \in \mathbb{R}^{d \times 4d}\), followed by a non-linear transformation using an activation function such as GELU. This is then followed by another linear transformation using \(\mathbf{W}^{\text{ffn}}_{\text{out}} \in \mathbb{R}^{4d \times d}\), as follows:

\vspace{-1.5em}

\begin{equation} \label{eqn:ffn_gelu}
\text{FFN}(\mathbf{X}) = (\text{GELU}(\mathbf{X} \mathbf{W}^{\text{ffn}}_{\text{in}}))\mathbf{W}^{\text{ffn}}_{\text{out}}
\end{equation}

\subsection{Entropy as a Metric for Attention Score Distribution} \label{Appendix:BasicOfEntropy}
Shannon's entropy quantifies the uncertainty in a probability distribution, measuring the amount of information needed to describe the state of a stochastic system \citep{shannon1948mathematical,jaynes1957information}. For a probability distribution \( P(x) \), the entropy is defined as \( \mathbf{E}(P) = -\sum_{i} P(x_i) \log P(x_i) \). Refer to \citep{baez2024entropy} for details on entropy. 

In a softmax-based attention mechanism, each softmax operation yields an entropy value representing the sharpness or spread of the attention scores for each query position \citep{ghader2017does,vig2019analyzing}. Higher entropy indicates a more uniform distribution of softmax scores, while lower entropy signifies a more focused distribution on certain features \citep{nahshan2024linear}.

Let \(\mathbf{A}^{(h,l)} \in \mathbb{R}^{T\times T} \) be the attention matrix of $h$-th head in $l$-th layer, and each element in the attention matrix, \( a_{ij}^{(l, h)} \), are attention weights, which are non-negative and sum to one for a query:

\vspace{-1.5em}

\begin{equation}
\mathbf{A}^{(l, h)} = \left[ a_{ij}^{(l, h)} \right]_{T \times T}, \quad \text{where} \quad a_{ij}^{(l, h)} \geq 0 \quad \text{and} \quad \sum_{j=1}^{T} a_{ij}^{(l, h)} = 1
\end{equation}

%\vspace{-1.5em}

This square matrix is generated by applying the softmax operation over the key length for each query position as follows (i.e., \(\mathbf{X} \in \mathbb{R}^{T\times T}\) \(\mathbf{X}_i \in \mathbb{R}^{1\times T}\) ): 

\vspace{-1.5em}

\begin{equation} 
\mathbf{A}^{(h,l)}(\mathbf{X}) = \text{Softmax}\Big(\frac{1}{\sqrt{d_k}} (\mathbf{X} \mathbf{W}^Q) (\mathbf{X}{\mathbf{W}^K})^\top \Big), \; \text{where} \quad \text{Softmax}(\mathbf{X}_i) = \frac{\exp \left(x_{i} \right)}{\sum_{j=1}^{T} \exp \left( x_{j} \right)}
\end{equation}

Thus, each element \( a_{ij}^{(l, h)} \) of the attention matrix can be represented as follows:

\vspace{-1.5em}

\begin{equation}
a_{ij}^{(l, h)} = \frac{\exp\left(\frac{1}{\sqrt{d_k}} (\mathbf{X}_i \mathbf{W}^Q) (\mathbf{X}_j \mathbf{W}^K)^\top \right)}{\sum_{k=1}^{T} \exp\left(\frac{1}{\sqrt{d_k}} (\mathbf{X}_i \mathbf{W}^Q) (\mathbf{X}_k \mathbf{W}^K)^\top \right).}
\end{equation}

Following \citep{zhai2023stabilizing}, we compute the mean of entropy values across all query positions to obtain a single entropy value for each head. The entropy \( \mathbf{E}^{(l, h)} \) for the \( h \)-th head in the \( l \)-th layer of an attention matrix is given by:

\vspace{-1.5em}

\begin{equation}
\mathbf{E}^{(l, h)} = -\frac{1}{T} \sum_{i=1}^{T} \sum_{j=1}^{T} a_{ij}^{(l, h)} \log \left( a_{ij}^{(l, h)} + \epsilon \right)
\end{equation}
where \( \epsilon \) is a small constant added for numerical stability to prevent taking the log of zero.

\begin{table}[t]
\caption{Architectural configurations of nonlinearities in LLMs, illustrating the combinations of Softmax (SM), LayerNorm (LN), GELU (G), and ReLU (R) functions (see Eq. \ref{eqn:ffn_mha}, \ref{eqn:attn_softmax}, \ref{eqn:mha_concat} and \ref{eqn:ffn_gelu}). } 
\label{tab:ArchConfigGPT2}
\centering 
\begin{tabular}{l|c} \toprule 
Abbreviation & Architectural configuration \\ \toprule 
\textcolor{blue}{SM} + \textcolor{violet}{LN} + \textcolor{red}{G} & $\mathbf{X}_{\text{out}} = \text{FFN}_{\text{\textcolor{red}{GELU}}}(\text{\textcolor{violet}{LayerNorm}}_{\textcolor{violet}{2}}(\text{MHA}(\text{Attn}_{\text{\textcolor{blue}{Softmax}}}(\text{\textcolor{violet}{LayerNorm}}_{\textcolor{violet}{1}}(\mathbf{X}_{\text{in}})))))$ \\
\textcolor{blue}{SM} + \textcolor{violet}{LN} + \textcolor{red}{R} & $\mathbf{X}_{\text{out}} = \text{FFN}_{\text{\textcolor{red}{ReLU}}}(\text{\textcolor{violet}{LayerNorm}}_{\textcolor{violet}{2}}(\text{MHA}(\text{Attn}_{\text{\textcolor{blue}{Softmax}}}(\text{\textcolor{violet}{LayerNorm}}_{\textcolor{violet}{1}}(\mathbf{X}_{\text{in}})))))$ \\ 
%\textcolor{blue}{SM} + \textcolor{violet}{LN} & $\mathbf{X}_{\text{out}} = \text{FFN}_{\text{Identity}}(\text{\textcolor{violet}{LayerNorm}}_{\textcolor{violet}{2}}(\text{MHA}(\text{Attn}_{\text{\textcolor{blue}{Softmax}}}(\text{\textcolor{violet}{LayerNorm}}_{\textcolor{violet}{1}}(\mathbf{X}_{\text{in}})))))$ \\ 
\textcolor{blue}{SM} + \textcolor{red}{G} & $\mathbf{X}_{\text{out}} = \text{FFN}_{\text{\textcolor{red}{GELU}}}(\text{MHA}(\text{Attn}_{\text{\textcolor{blue}{Softmax}}}(\mathbf{X}_{\text{in}})))$ \\ 
\textcolor{blue}{SM} + \textcolor{red}{R} & $\mathbf{X}_{\text{out}} = \text{FFN}_{\text{\textcolor{red}{ReLU}}}(\text{MHA}(\text{Attn}_{\text{\textcolor{blue}{Softmax}}}(\mathbf{X}_{\text{in}})))$ \\ 
%\textcolor{blue}{SM} & $\mathbf{X}_{\text{out}} = \text{FFN}_{\text{Identity}}(\text{MHA}(\text{Attn}_{\text{\textcolor{blue}{Softmax}}}(\mathbf{X}_{\text{in}})))$ \\ 
\bottomrule
\end{tabular} 
\end{table}

The combination of MHA and FFN sub-blocks, along with residual connections and normalization layers, allows transformer models to  learn the contextual relationships between tokens effectively.

\subsection{Dataset and Training Methodology}

We train our models from scratch using the CodeParrot dataset \cite{codeParrot}, a standard benchmark for LLMs \citep{he2024simplifying,he2024understanding}. The dataset, derived from 20 million Python files on GitHub, consists of 8 GB of data with 16.7 million examples, each containing 128 tokens, amounting to a total of 2.1 billion training tokens. For tokenization, we utilize a tokenizer with a vocabulary size of 50K. 

For training on the CodeParrot dataset, we adopt the settings from \citep{he2024simplifying}, ensuring consistency across all architectural variations to isolate the effects of the changes. In line with prior works \cite{he2024simplifying,stanic2023languini,geiping2023cramming}, all models are trained using a single RTX 3090 GPU.
\section{Activation Functions and Their Impact Through Shannon's Entropy}
In this section, we investigate the role of activation functions in baseline and normalization-free  decoder-only LLMs. Specifically, we examine the learning dynamics and internal representations of activation functions, using entropy as a metric to highlight key observations and insights.

\begin{figure} [t]
\centering
\subfloat[Headwise entropy distribution \label{subfig:EntropyBars}]{\includegraphics[width=.49\textwidth]{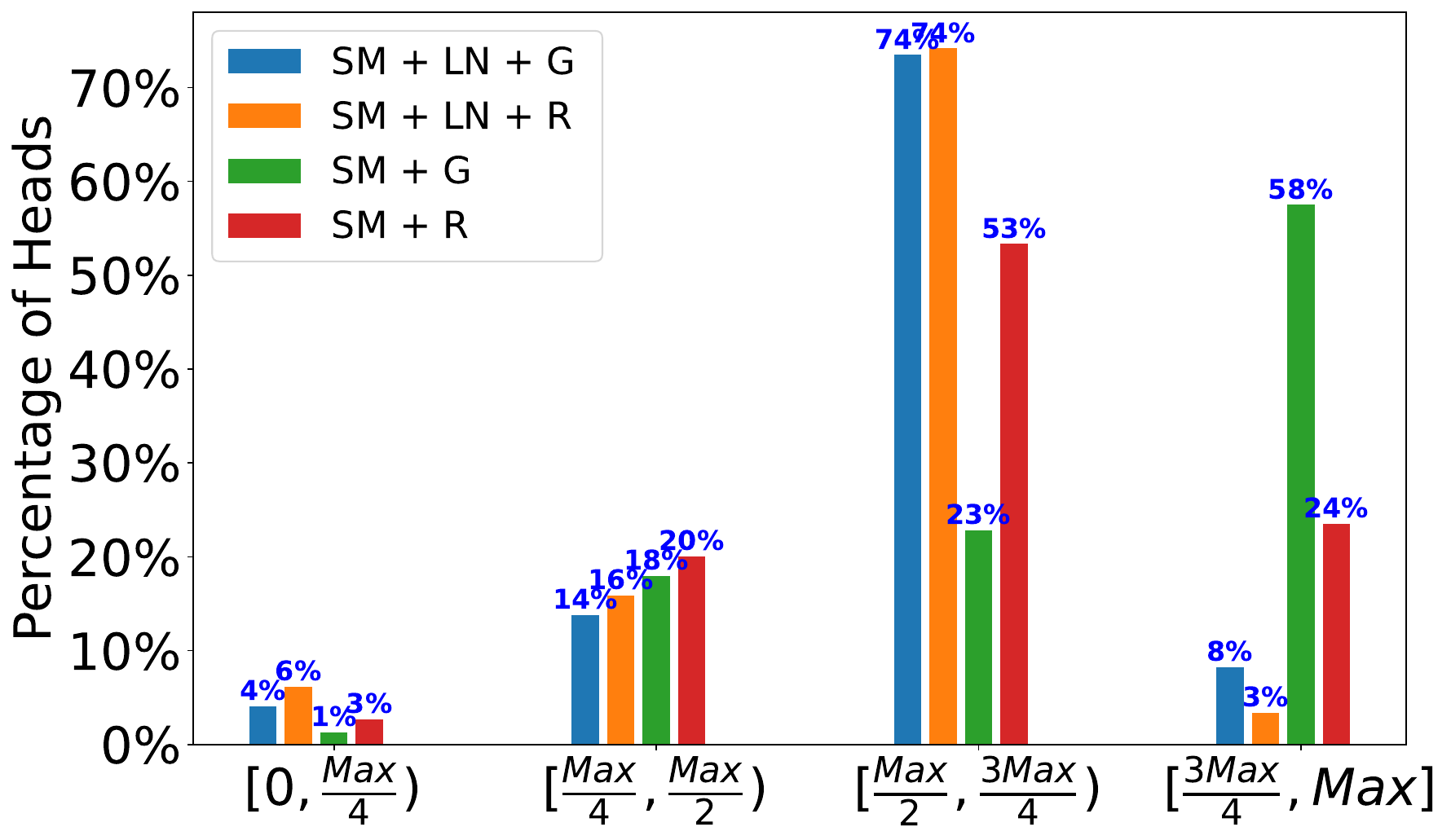}} 
\subfloat[Evaluation loss curve]{\includegraphics[width=.49\textwidth]{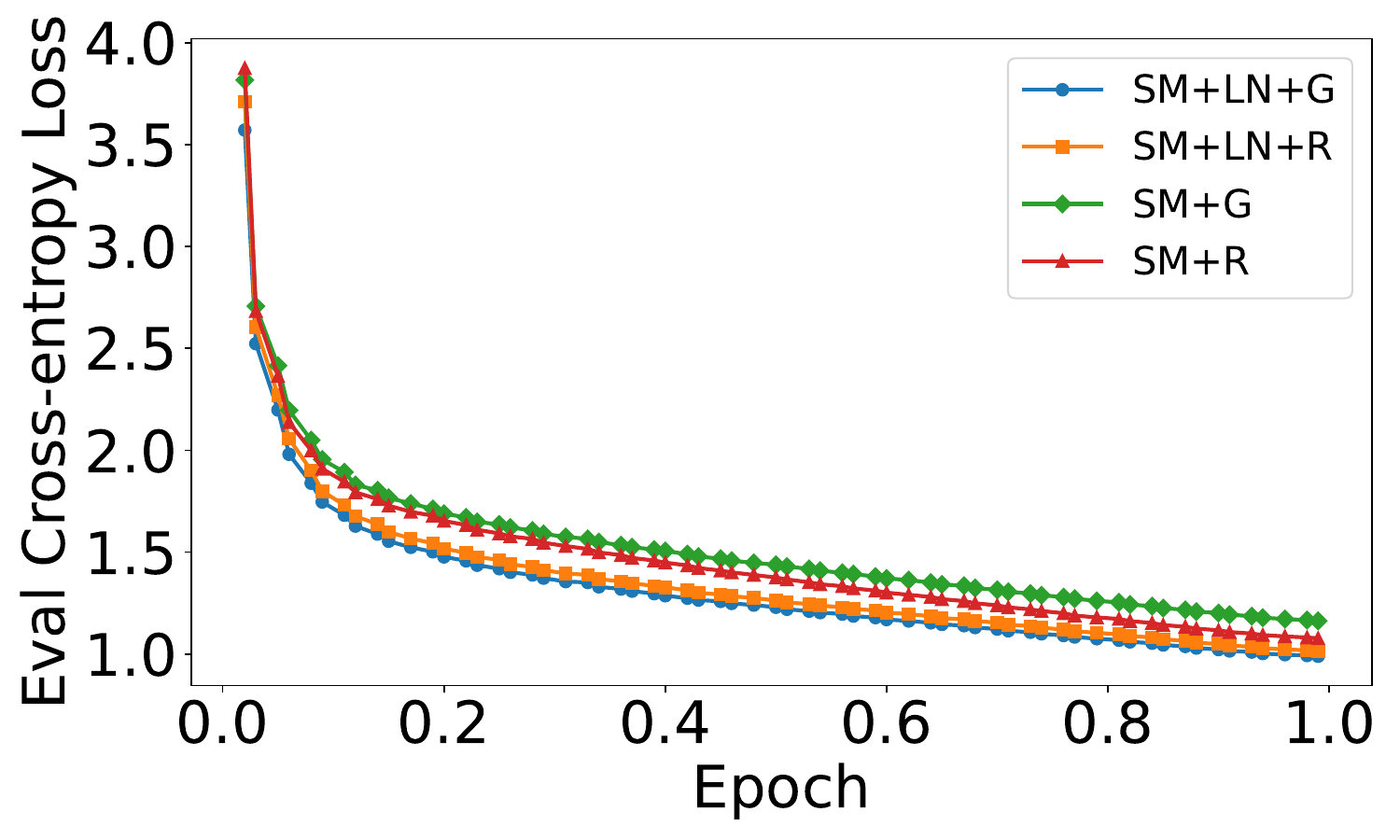}} \vspace{-0.5em}
\caption{Headwise entropy distribution and evaluation loss for baseline and normalization-free GPT-2 models, using GELU and ReLU activations, trained from scratch on CodeParrot dataset.} 
\label{fig:LossCurveNonlinConfig}
\end{figure}

{\bf Well-behaved entropy distribution} 
We begin by analyzing the headwise entropy distribution of baseline architecture with GELU and ReLU in the FFN,  i.e., configurations ${\tt SM + LN + G}$ and ${\tt SM + LN + R}$ respectively. We find that the majority of heads ($\approx$90\%) possess entropy values between $\frac{\text{max}}{4}$ and $\frac{\text{3max}}{4}$, where ${\tt max}$ is maximum observed entropy value among all heads (Figure \ref{subfig:EntropyBars}).  This concentration in the mid-entropy range, avoiding extremes, demonstrates a well-behaved distribution, providing as a benchmark for assessing the impact of architectural modifications, such as activation function simplification, on model behavior.

{\bf Entropic overload}
We observed that in certain nonlinearity configurations, a disproportionately large fraction of the attention heads exhibit higher entropy values (between $\frac{3\text{max}}{4}$ and ${\tt max}$). We term this phenomenon as entropic overload and hypothesize that this imbalance results in {\em under-utilization} of the network's representational capacity, as too many heads engaged in exploration, hindering the model from effectively leveraging the diversity of attention heads.

\begin{figure} [htbp]
\centering
\subfloat[SM + LN + G \label{subfig:sm_ln_g}]{\includegraphics[width=.5\textwidth]{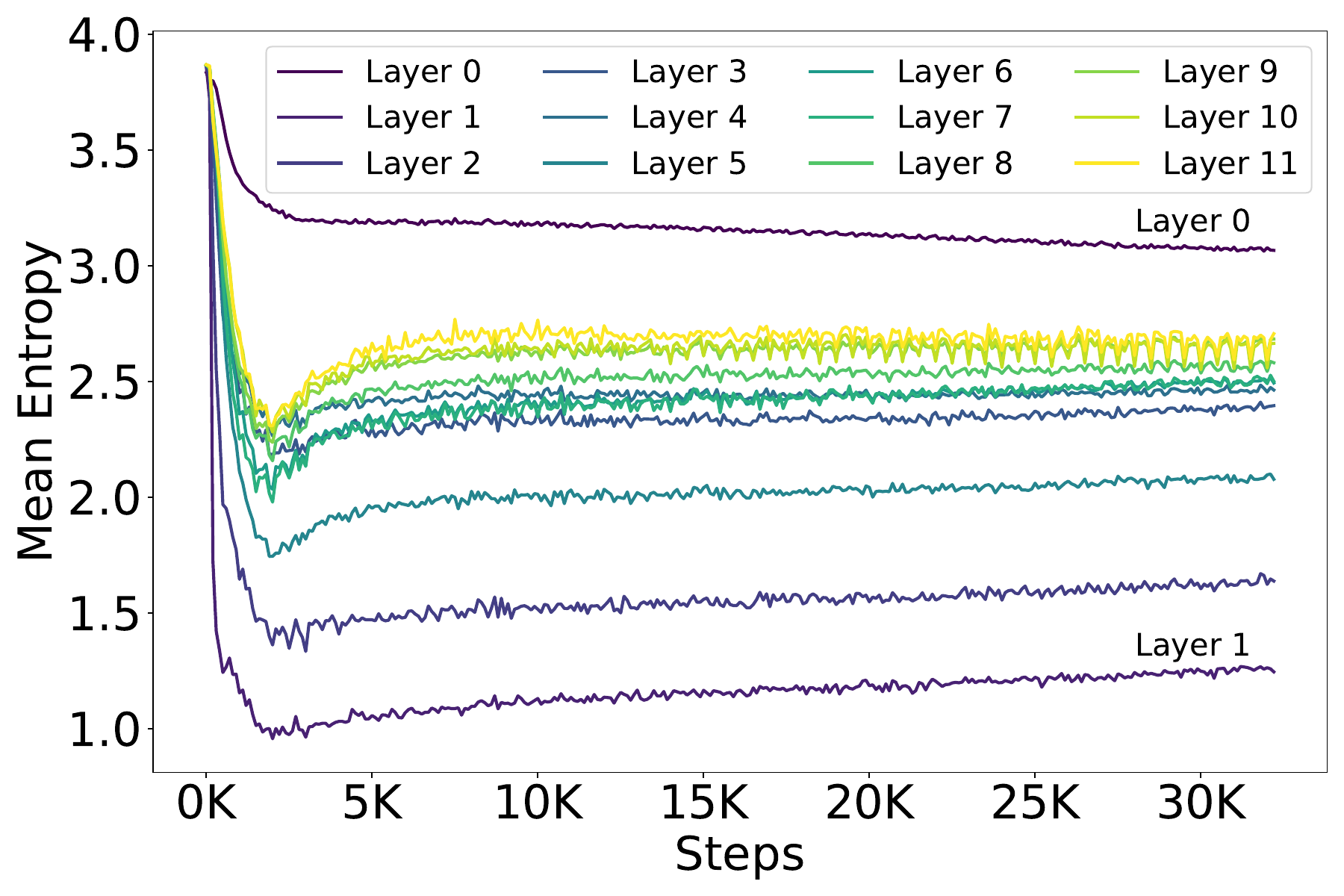}} 
\subfloat[SM + LN + R \label{subfig:sm_ln_r}]{\includegraphics[width=.5\textwidth]{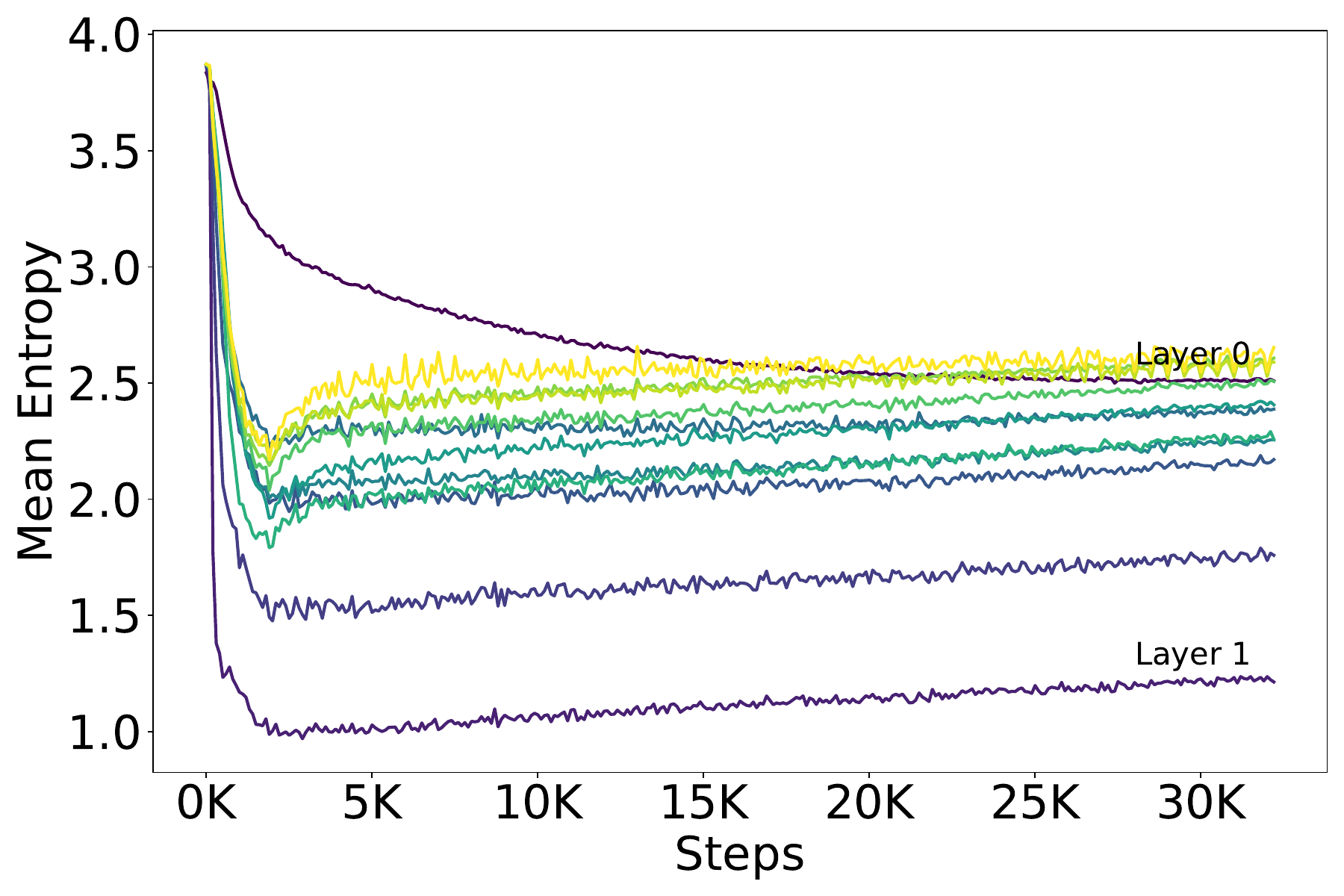}} \\ \vspace{-0.8em}
\subfloat[SM + G]{\includegraphics[width=.5\textwidth]{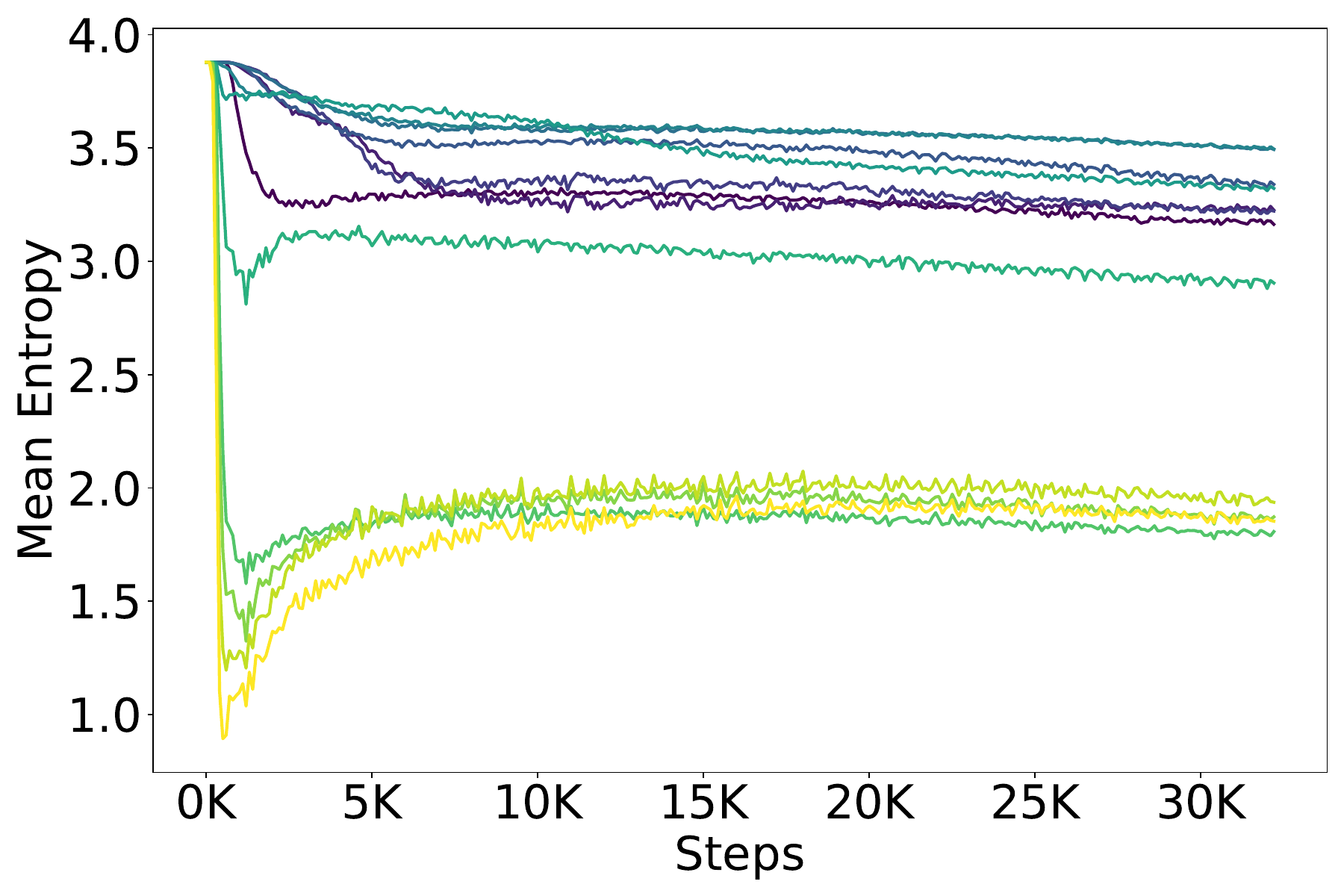}} 
\subfloat[SM + R \label{subfig:sm_r}]{\includegraphics[width=.5\textwidth]{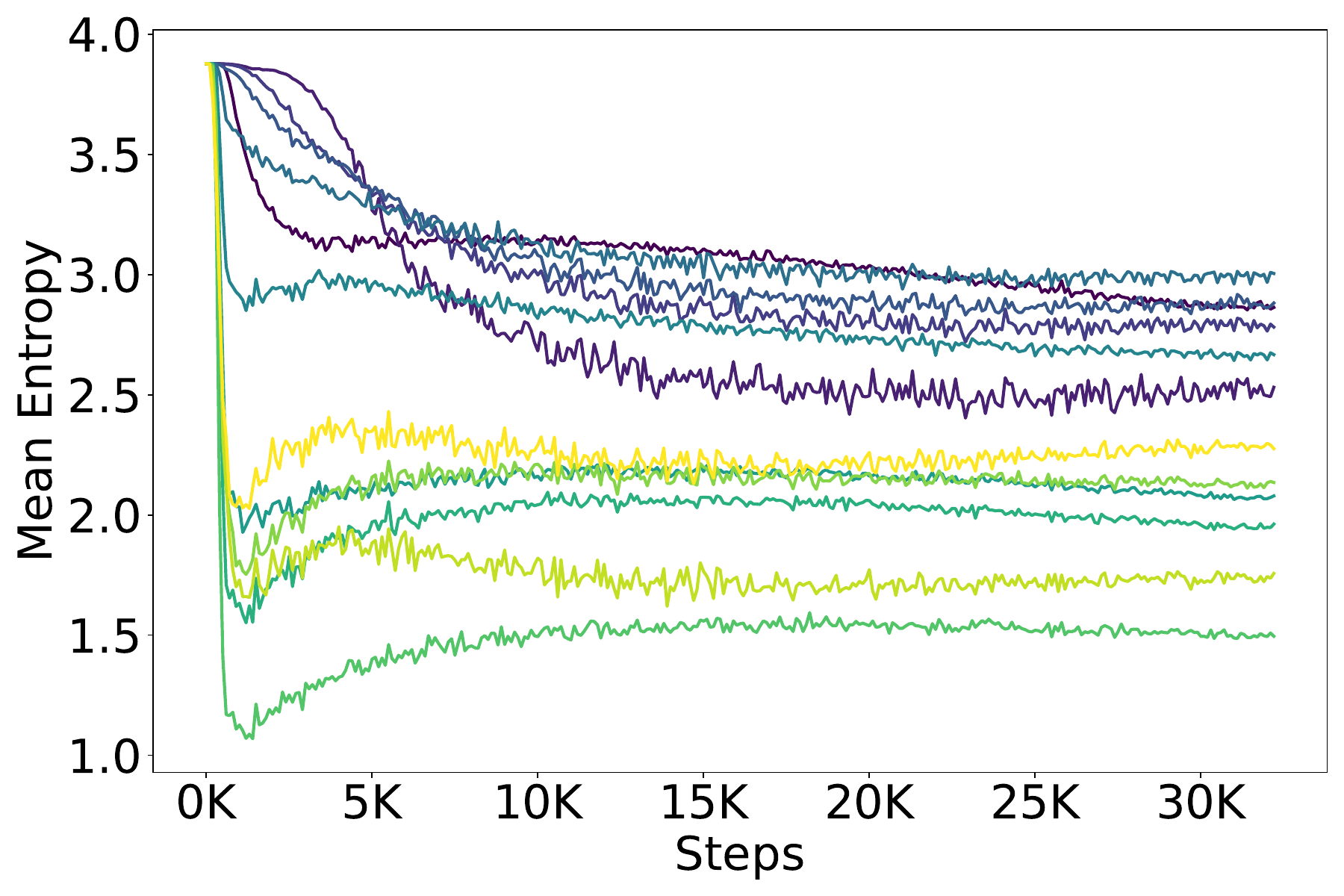}} \\ \vspace{-0.8em}
\subfloat[SM + LearnableNegativeSlope (Layerwise) \label{subfig:LeakyLayerwise}]{\includegraphics[width=.5\textwidth]{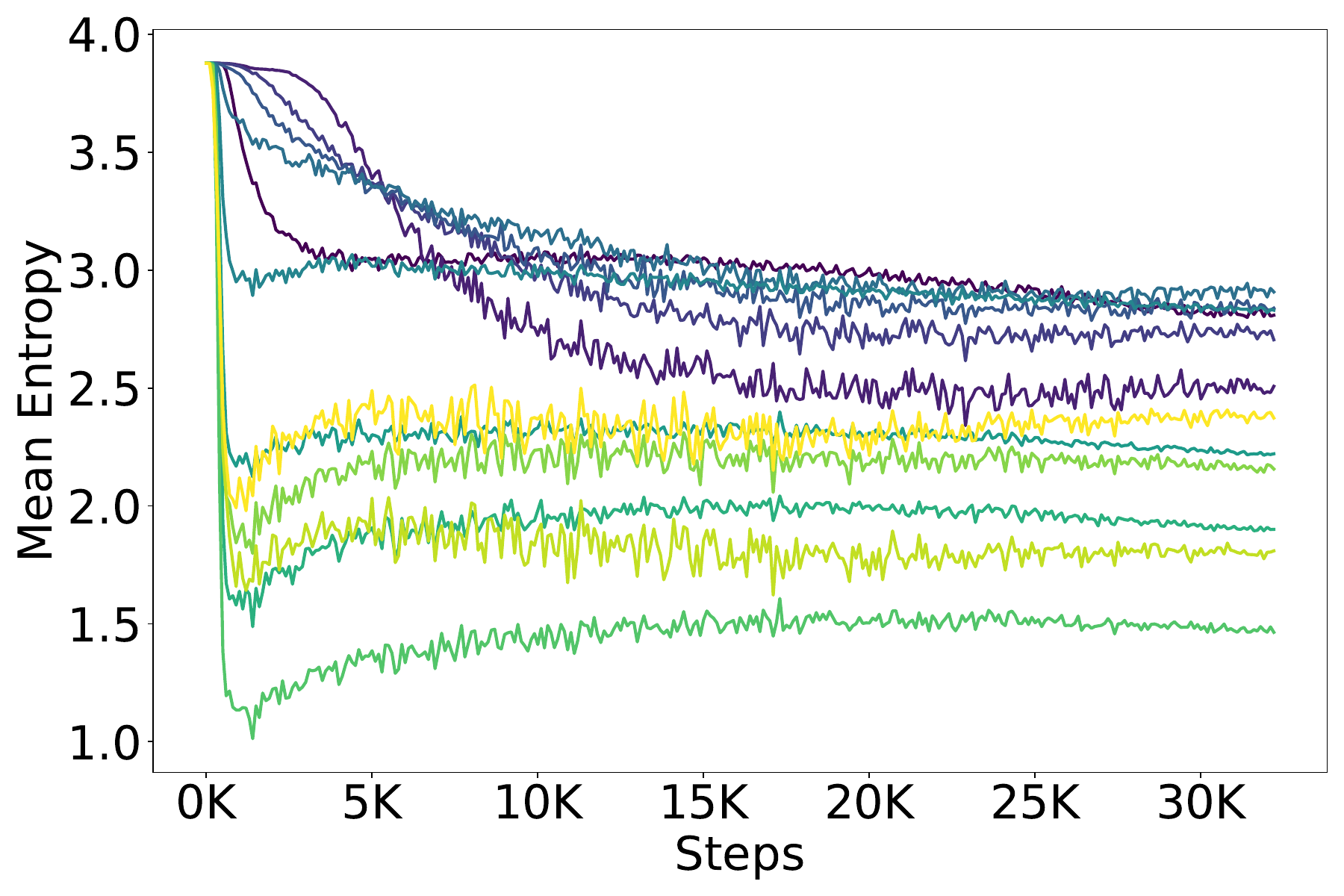}} 
\subfloat[SM + LearnableNegativeSlope (Global) \label{subfig:LeakyGlobal}]{\includegraphics[width=.5\textwidth]{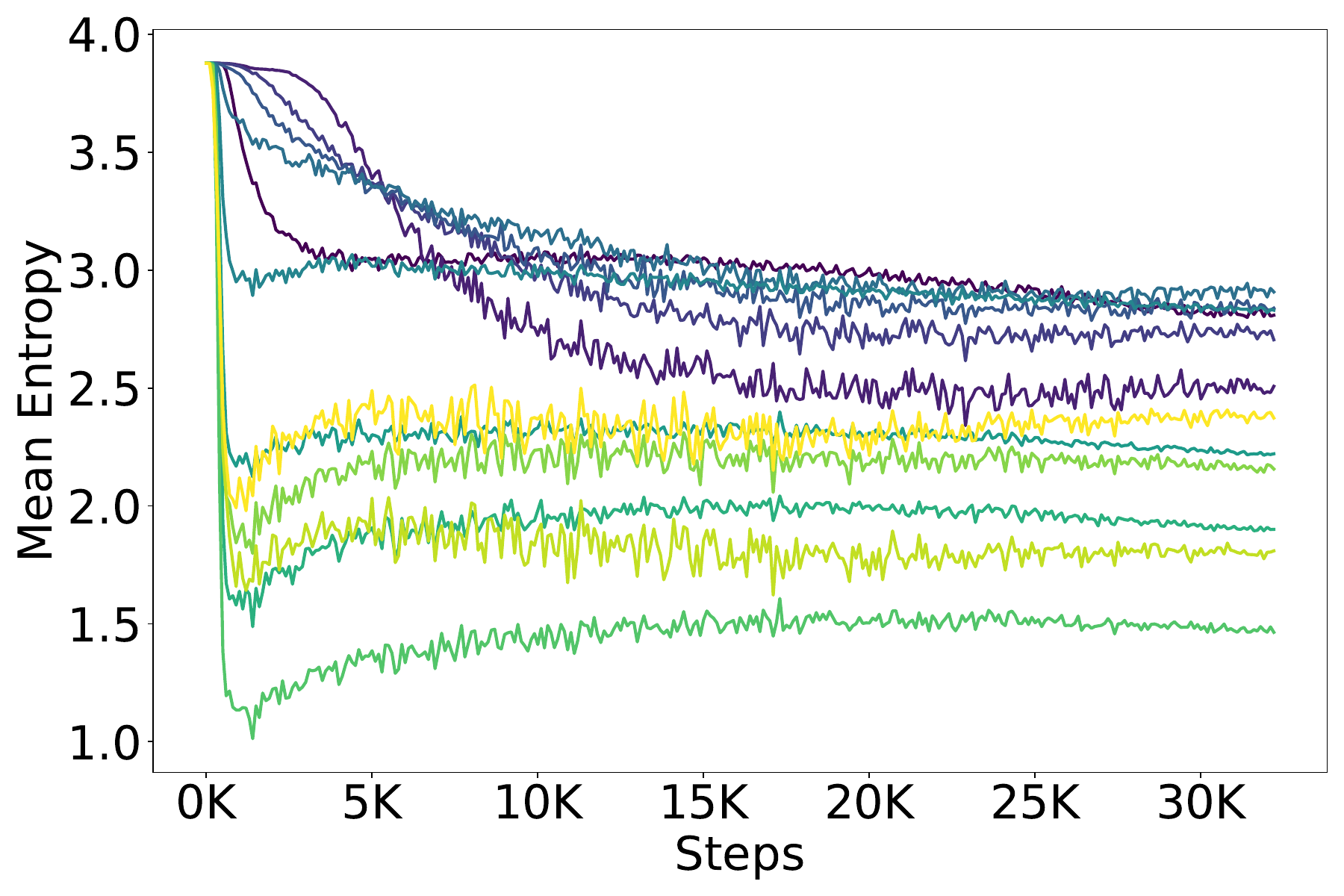}} 
\vspace{-0.2em}
\caption{Evolution of Layerwise entropy when GPT-2 ($L$=12, $H$=12, $d$=768) models with various nonlinearity configurations are trained from scratch on CodeParrot dataset. Evolution of layer-wise entropy during training of GPT-2 models ($L$=12, $H$=12, $d$=768) with different nonlinearity configurations on the CodeParrot dataset. The near-identical entropy dynamics in Figures d, e, and f underscore a {\em natural preference} for a zero negative slope, similar to ReLU, in the FFN activation function of the normalization-free model. } 
\label{fig:LayerwiseEntropy}
\end{figure}

To investigate further, we examined how entropy values evolve during training. Typically, all heads start with higher entropy values, indicating an initial exploration phase, and gradually adapt to balance exploration and exploitation in baseline networks (see Figure \ref{fig:LayerwiseEntropy}). However, in the absence of certain nonlinearities, this balance is disrupted, preventing attention heads from specializing and refining their focus on critical aspects of the input, thereby diminishing overall performance.

%Entropy distribution across various configurations of nonlinearities in GPT-2.

%\begin{table}[t]
%\begin{minipage}[b]{0.56\linewidth}
%\centering
%\includegraphics[width=.99\textwidth]{plots/plot_loss_curve.pdf}  
%\captionof{figure}{Evaluation loss for baseline and normalization-free GPT-2 models with GELU and ReLU activations.}
%\label{fig:LossCurveNonlinConfig}
%\end{minipage}  \hfill
%\begin{minipage}[b]{0.43\linewidth}
%\centering 
%\resizebox{0.99\textwidth}{!}{
%\begin{tabular}{lcl} \toprule
%Configurations & Eval PPL & +$\Delta$(\%)\\ \toprule
%SM + LN + G & 2.69 & 0.00  \\
%SM + LN + R & 2.76 & 2.53  \\
%SM + G & 3.20 &      18.92     \\
%SM + R & 2.94 &      9.20      \\  \bottomrule 
%\end{tabular} }  
%\captionof{table}{Perplexity comparison of baseline and normalization-free GPT-2 models with GELU and ReLU activations in the FFN, trained from scratch. While the baseline model with GELU outperforms its ReLU counterpart, the normalization-free design shows the {\em opposite trend}.}
%%\label{tab:NonlinConfigEvalPPL}
%\end{minipage}
%\end{table}

%\vspace{-1em}

{\bf Observation 1: ReLU significantly outperforms GELU in LayerNorm-Free LLMs.} 
While GELU is typically preferred over ReLU in conventional transformer-based models due to its smooth and differentiable properties that improve performance and optimization, our empirical findings indicate the {\em opposite trend} for LayerNorm-free models--- using ReLU in the FFN exhibit better learning dynamics than their GELU counterpart. This leads to an {\bf 8.2\%} improvement in perplexity for GPT-2 (see Figure \ref{fig:LossCurveNonlinConfig} and Table \ref{tab:NonlinConfigEvalPPL}). A similar trend has been observed on the normalization-free Pythia-70M model across various context lengths.  

\vspace{-1.5em}

\begin{table}[htbp]
\caption{Perplexity comparison between baseline and normalization-free GPT-2 ($L$=12, $H$=12, $d$=768) and Pythia-70M ($L$=6, $H$=8, $d$=512) models, using GELU and ReLU activations in the FFN, trained from scratch on CodeParrot dataset. While GELU outperforms ReLU in baseline models, the normalization-free models exhibit the {\em opposite trend}. }
\label{tab:NonlinConfigEvalPPL}
\centering
\begin{tabular}{lcccccc} \toprule
& \multicolumn{2}{c}{GPT-2 ($T$=128)} & \multicolumn{2}{c}{Pythia-70M ($T$=128)} & \multicolumn{2}{c}{Pythia-70M ($T$=256)} \\
\cmidrule(lr){2-3} \cmidrule(lr){4-5} \cmidrule(lr){6-7}
& Eval PPL &+$\Delta$(\%) & Eval PPL &+$\Delta$(\%) & Eval PPL & +$\Delta$(\%) \\ \toprule
SM+LN+G & 2.688 & 0.00 & 3.512 & 0.00 & 3.054 & 0.00 \\
SM+LN+R & 2.757 & 2.53 & 3.590 & 2.22 & 3.107 & 1.73 \\
SM+G & 3.197 & 18.92 & 4.086 & 16.35 & 3.570 & 16.87 \\
SM+R & 2.936 & 9.20 & 3.736 & 6.36 & 3.273 & 7.17 \\
\bottomrule
\end{tabular} 
\end{table}

\begin{figure} [htbp]
\centering
\subfloat[Layerwise learnable slope]{\includegraphics[width=.5\textwidth]{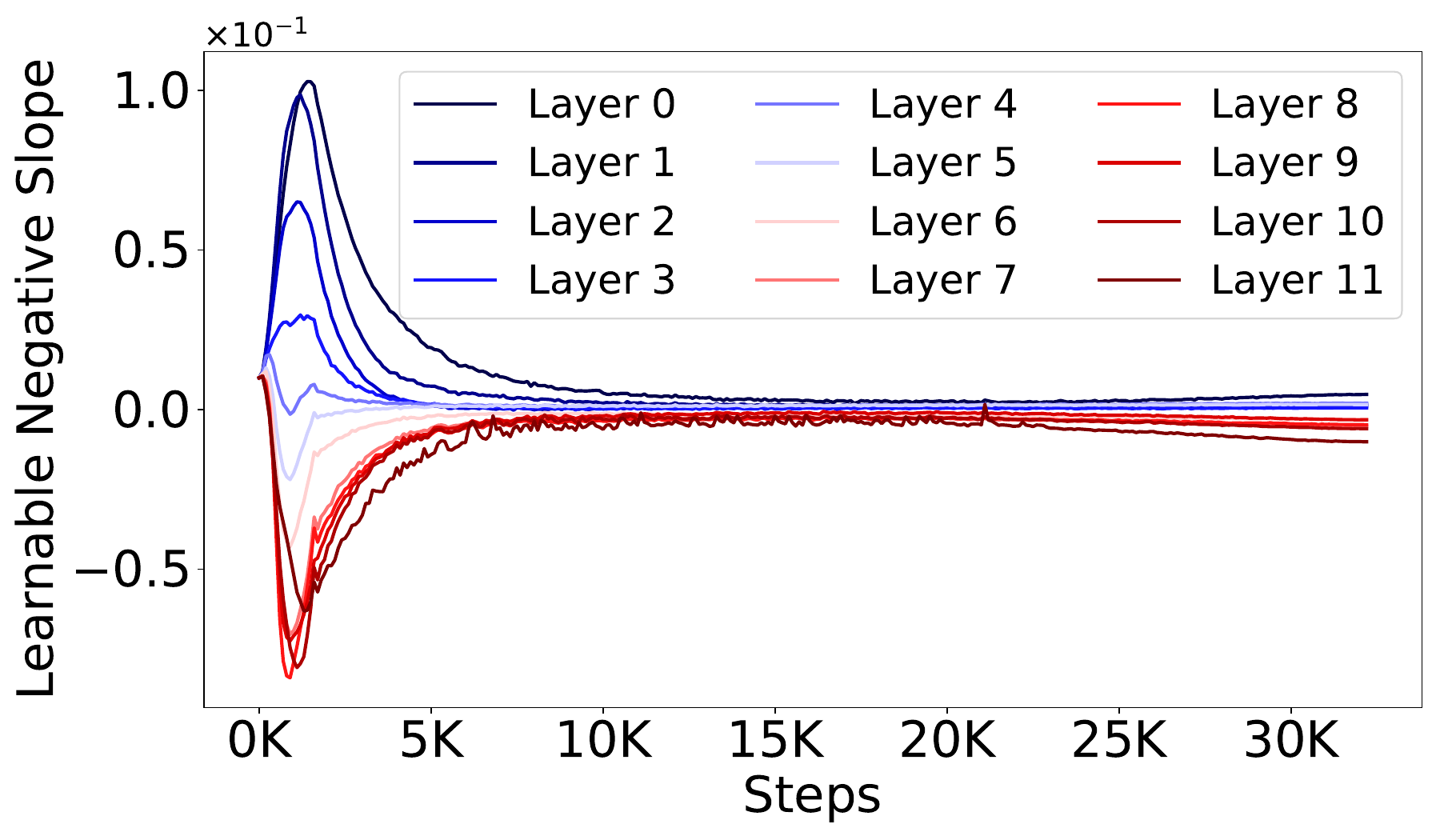}} 
\subfloat[Global learnable slope]{\includegraphics[width=.5\textwidth]{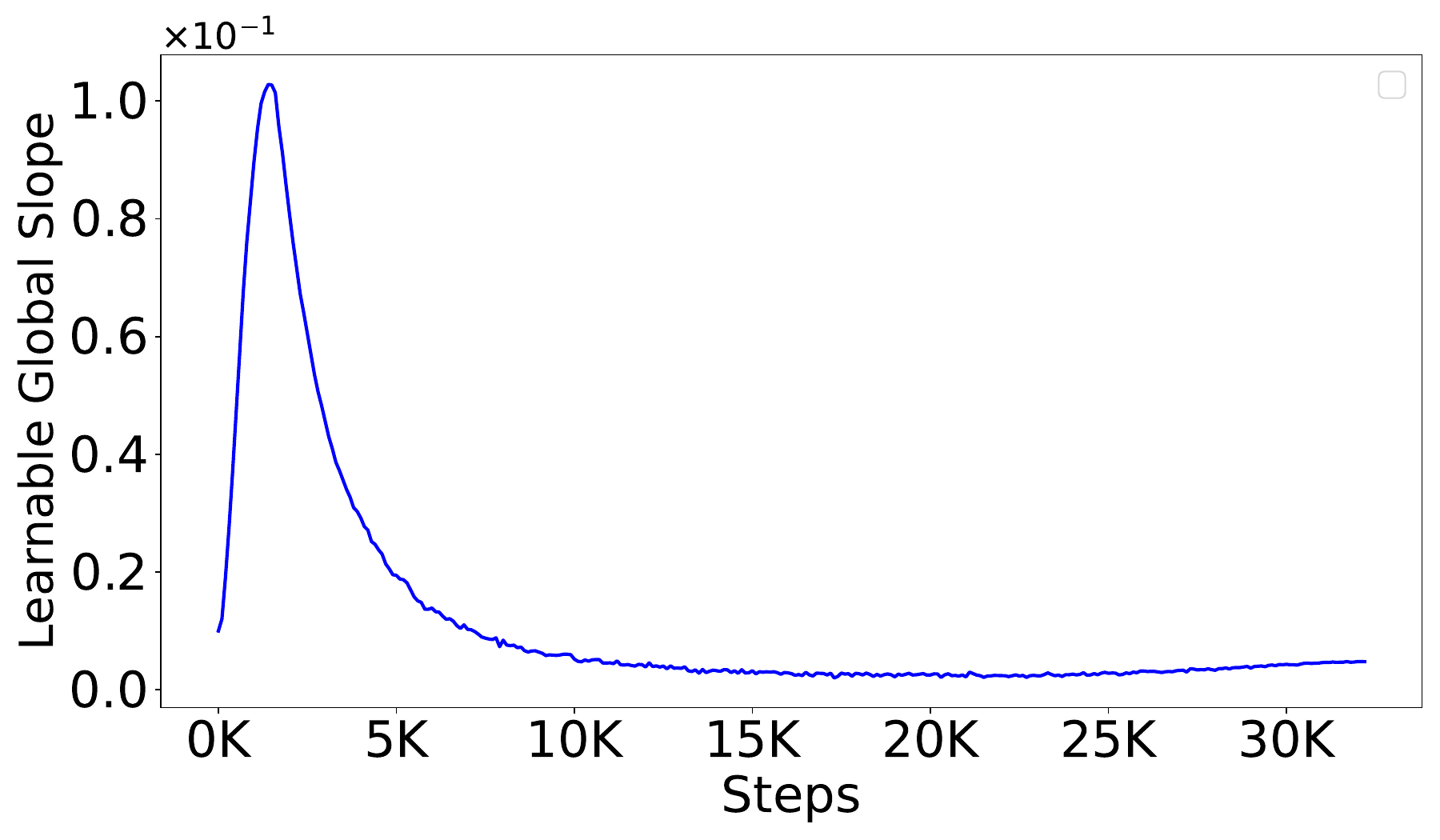}} \vspace{-0.5em}
%\subfloat[Value of learnable scaling factors]{\includegraphics[width=.33\textwidth]{plots/inverse_mlp_block_resid_gain}}  
\caption{Learnable negative slope for leaky ReLU in FFN of LN-free GPT-2 model. (a) Layerwise slopes showing initial variability and convergence towards zero. (b) Global slope trend towards zero over training steps, indicating a preference for zero negative slope in LN-free architectures. } 
\label{fig:LearnableNegSlope} 
\end{figure}

To further strengthen our findings, we conducted experiments with a learnable negative slope in the leaky ReLU activation function with two configurations: 1) layer-wise, where each layer has its independent learnable slope, and 2) global, where a single learnable slope is shared across all layers. Interestingly, in the layerwise setting, the early layers initially learn a positive slope while the deeper layers learn a negative slope. However, as training progresses, all layers converge to a near-zero slope. In the global setting, the slope first shifts to positive before converging to near zero (see Figure \ref{fig:LearnableNegSlope}). 

\vspace{-1.5em}

When comparing the layerwise entropy dynamics in both cases (Figure \ref{subfig:LeakyLayerwise} and Figure \ref{subfig:LeakyGlobal}) with the normalization-free model using ReLU activations (Figure \ref{subfig:sm_r}), we observed near-identical patterns. This highlights the a {\em natural preference} for zero negative slope, similar to ReLU, in the FFN activation function of the normalization-free model.

{\bf Observation 2: Early layers in the LayerNorm-Free model with GELU in FFN experience entropic overload}. 
To understand the zero negative slope preference for the FFN activation function in LN-free architecture, we analyzed the headwise entropy values of LN-free models with GELU and ReLU, when trained from scratch, and compared them to their baseline counterparts. Our analysis revealed a significant divergence in the headwise entropy distributions of the LN-free GELU model (see Figure \ref{fig:AttnEntHeatMaps}). While baseline models with GELU and ReLU exhibit a balanced entropy distribution, by avoiding the extreme values,  the LN-free GELU model shows entropic overload in early layers.

Quantitatively, 58\% of heads in the LN-free GELU model have entropy values between $\frac{\text{3max}}{4}$ and ${\tt max}$, compared to only 23\% in the LN-free ReLU model (Figure \ref{fig:LossCurveNonlinConfig}a). More importantly, very few heads in the latter approach maximum entropy compared to the former (see yellow regions in Figure \ref{fig:AttnEntHeatMaps}c), indicating more severe entropic overload in the LN-free model with GELU.

These observations align with geometrical properties of ReLUs: it preserve more information about the structure of the raw input, encouraging neurons to specialize in different regions of the input space,  leading to a higher intra-class {\em selectivity} and {\em specialization} \citep{alleman2024task}. Thus, the lack of LayerNorm makes the geometry and specialization effects of ReLU more beneficial, while GELU's smoother nonlinearity causes issues in maintaining distinct attention head behaviors.

\begin{figure} [htbp]
\centering
\subfloat[NaNs observed with a negative slope of $1e$-2 \label{subfig:NaNs1e2}]{\includegraphics[width=.5\textwidth]{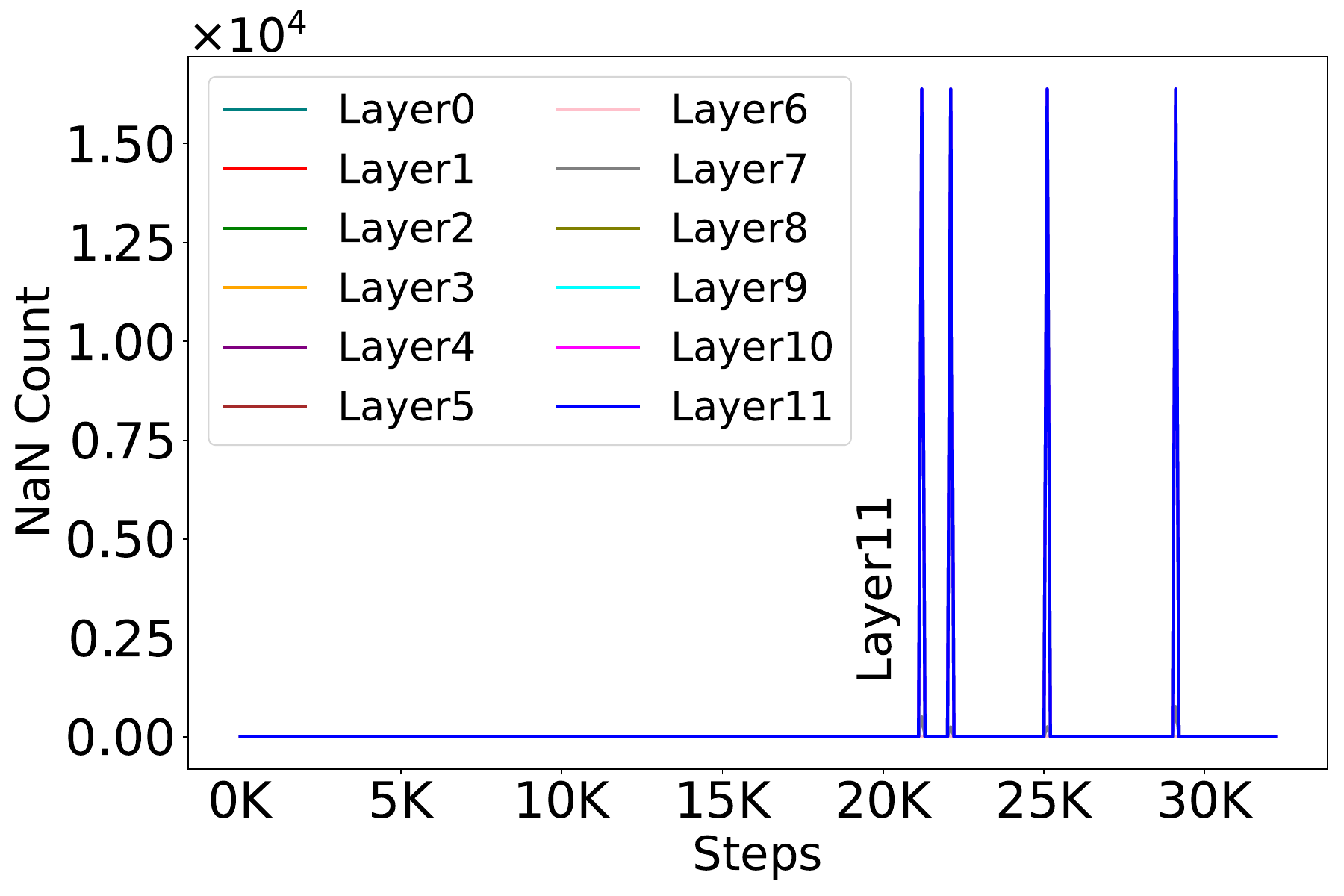}} 
\subfloat[Entropy dynamics with a negative slope of $1e$-2 \label{subfig:Ent1e2}]{\includegraphics[width=.48\textwidth]{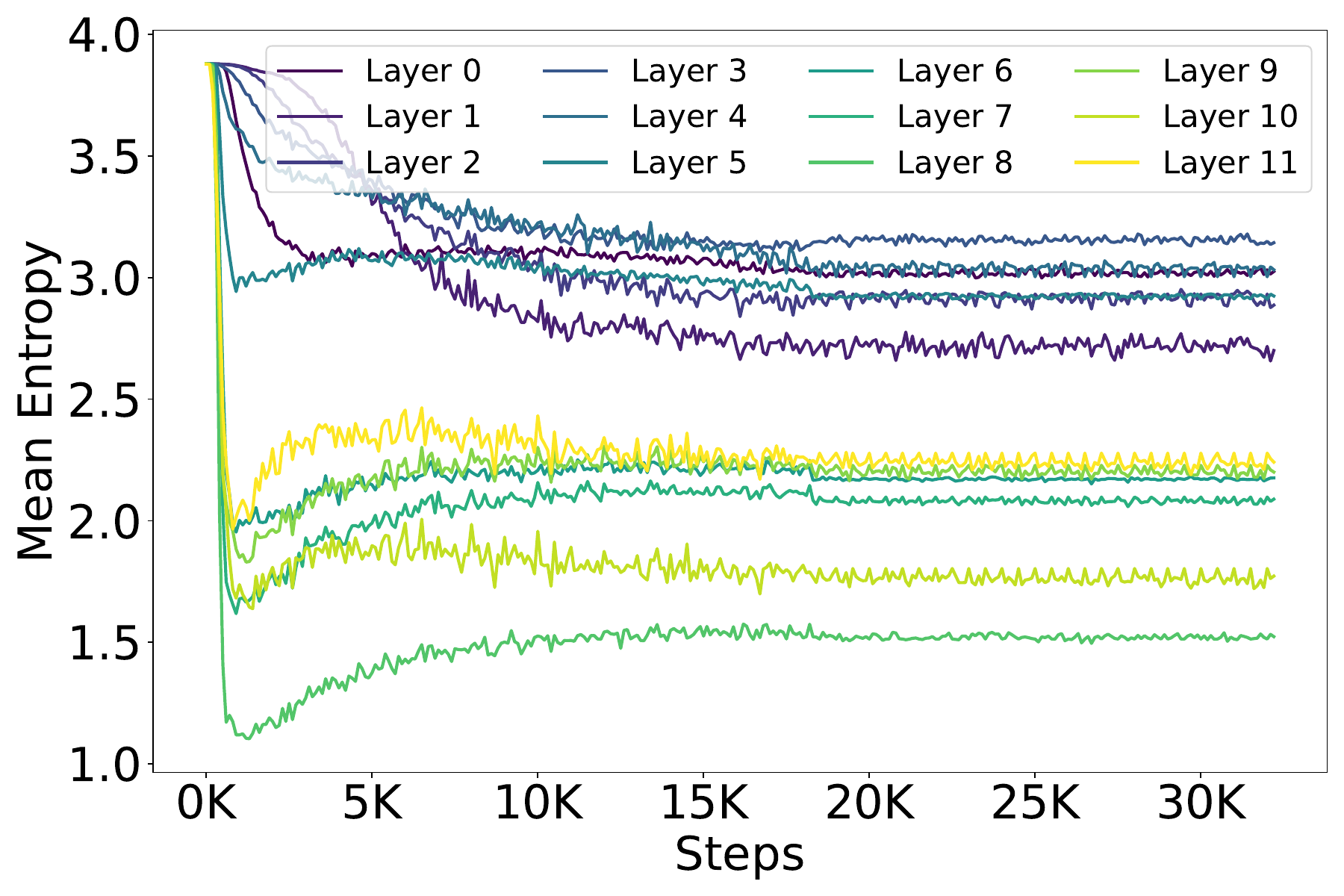}} \\ \vspace{-0.8em}
\subfloat[NaNs observed with a negative slope of $5e$-2 \label{subfig:NaNs5e2}]{\includegraphics[width=.5\textwidth]{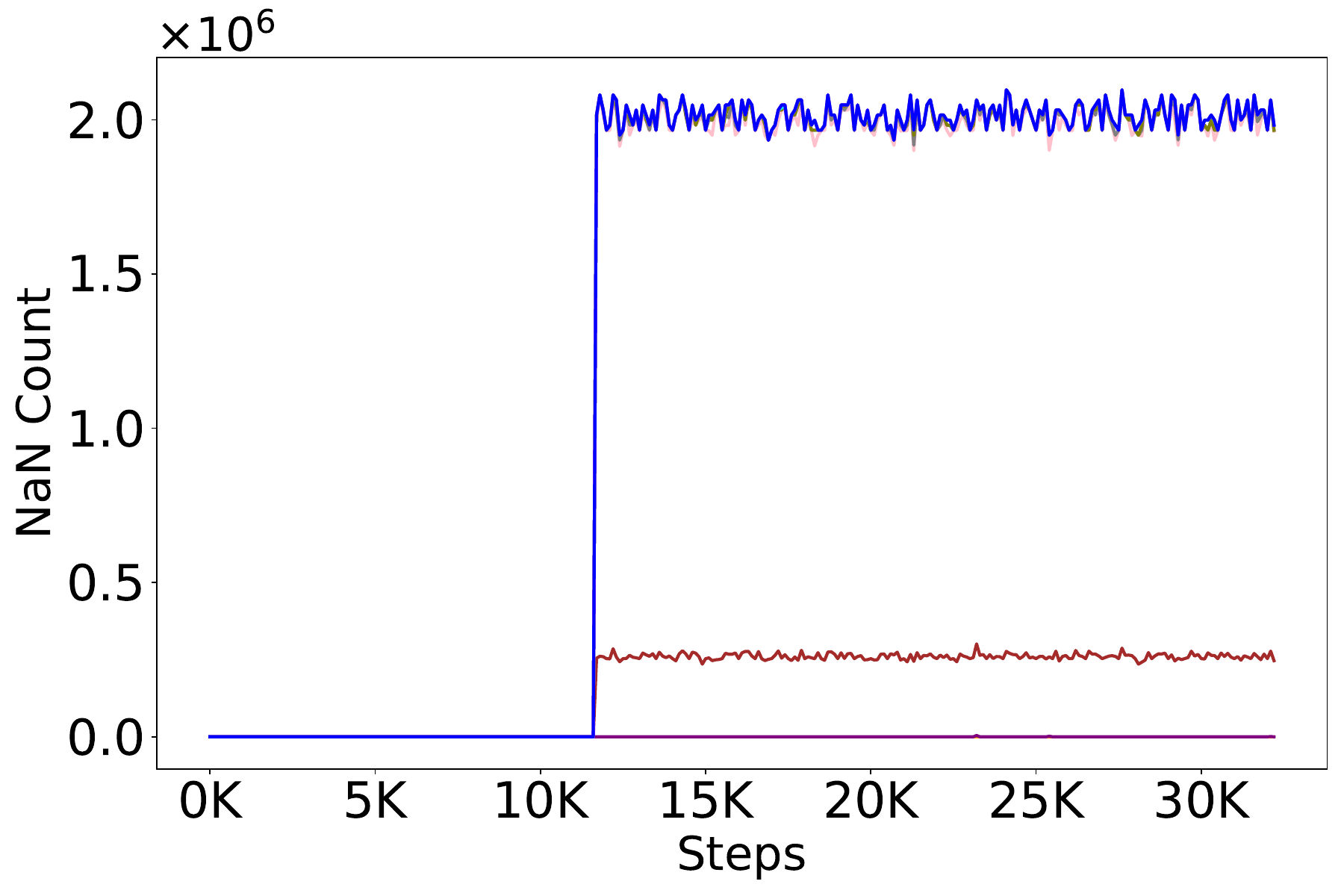}} 
\subfloat[Entropy dynamics with a negative slope of $5e$-2 \label{subfig:Ent5e2}]{\includegraphics[width=.48\textwidth]{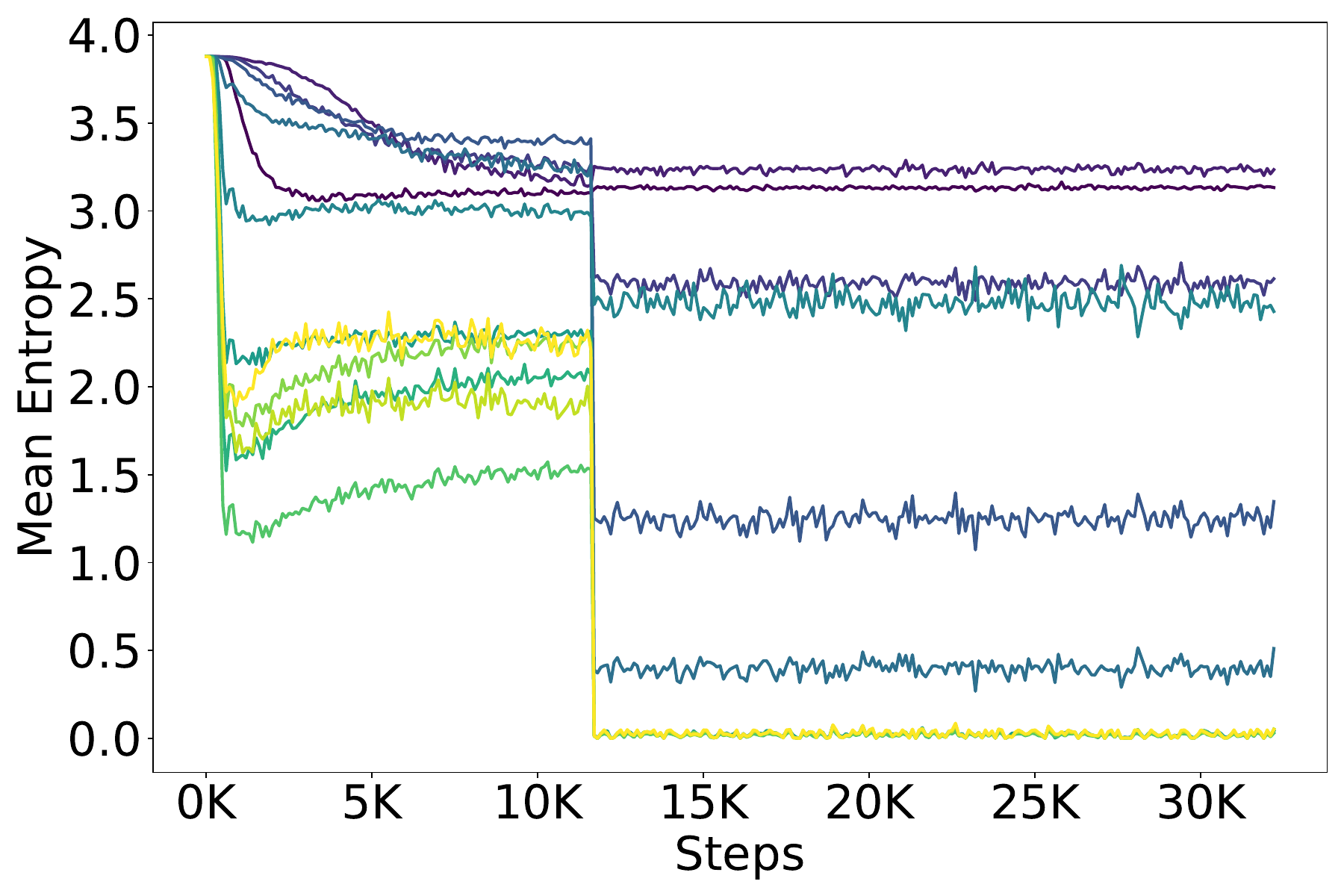}} \\ \vspace{-0.8em}
\subfloat[NaNs observed with a negative slope of $1e$-1 \label{subfig:NaNs1e1}]{\includegraphics[width=.5\textwidth]{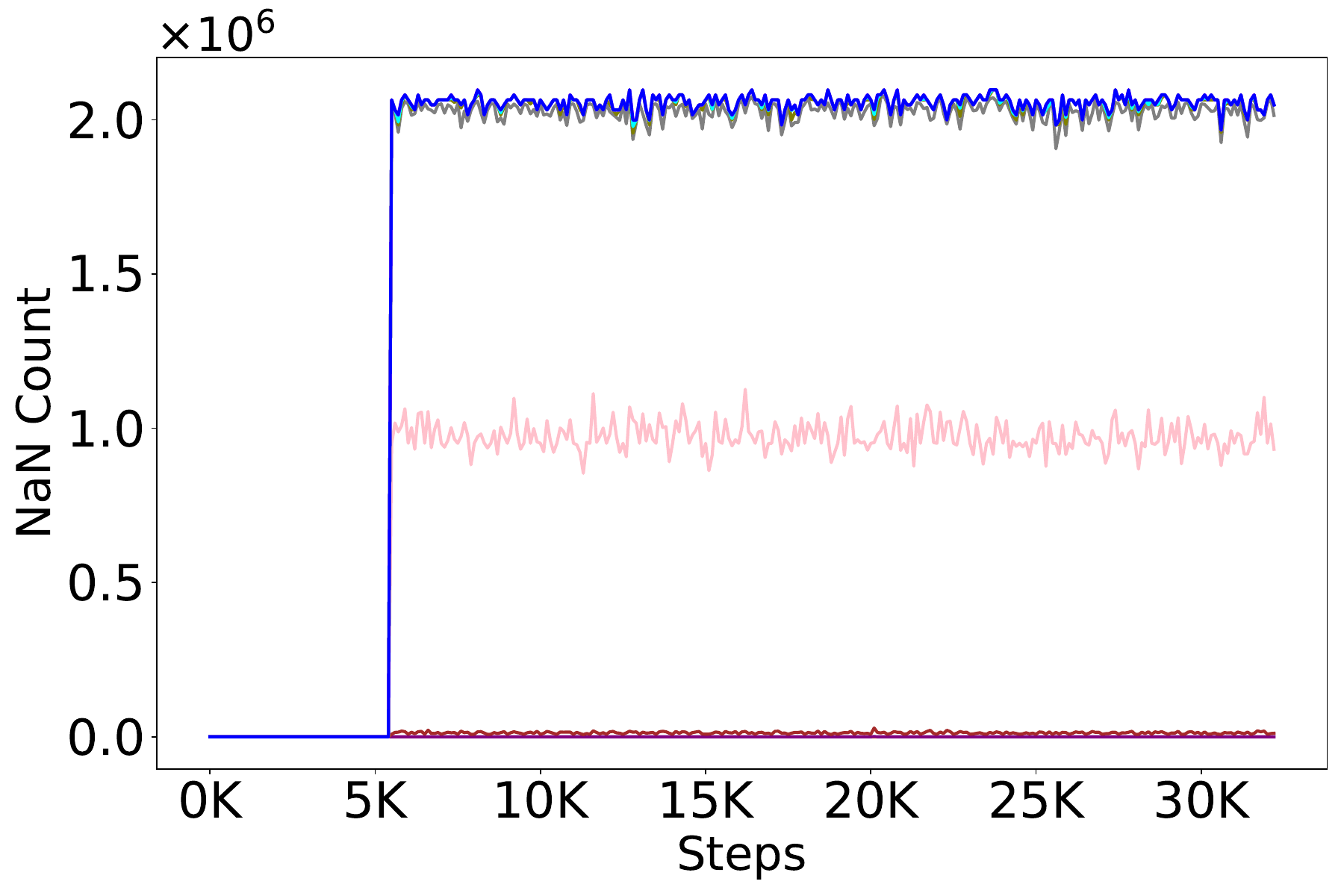}} 
\subfloat[Entropy dynamics with a negative slope of $1e$-1 \label{subfig:Ent1e1}]{\includegraphics[width=.48\textwidth]{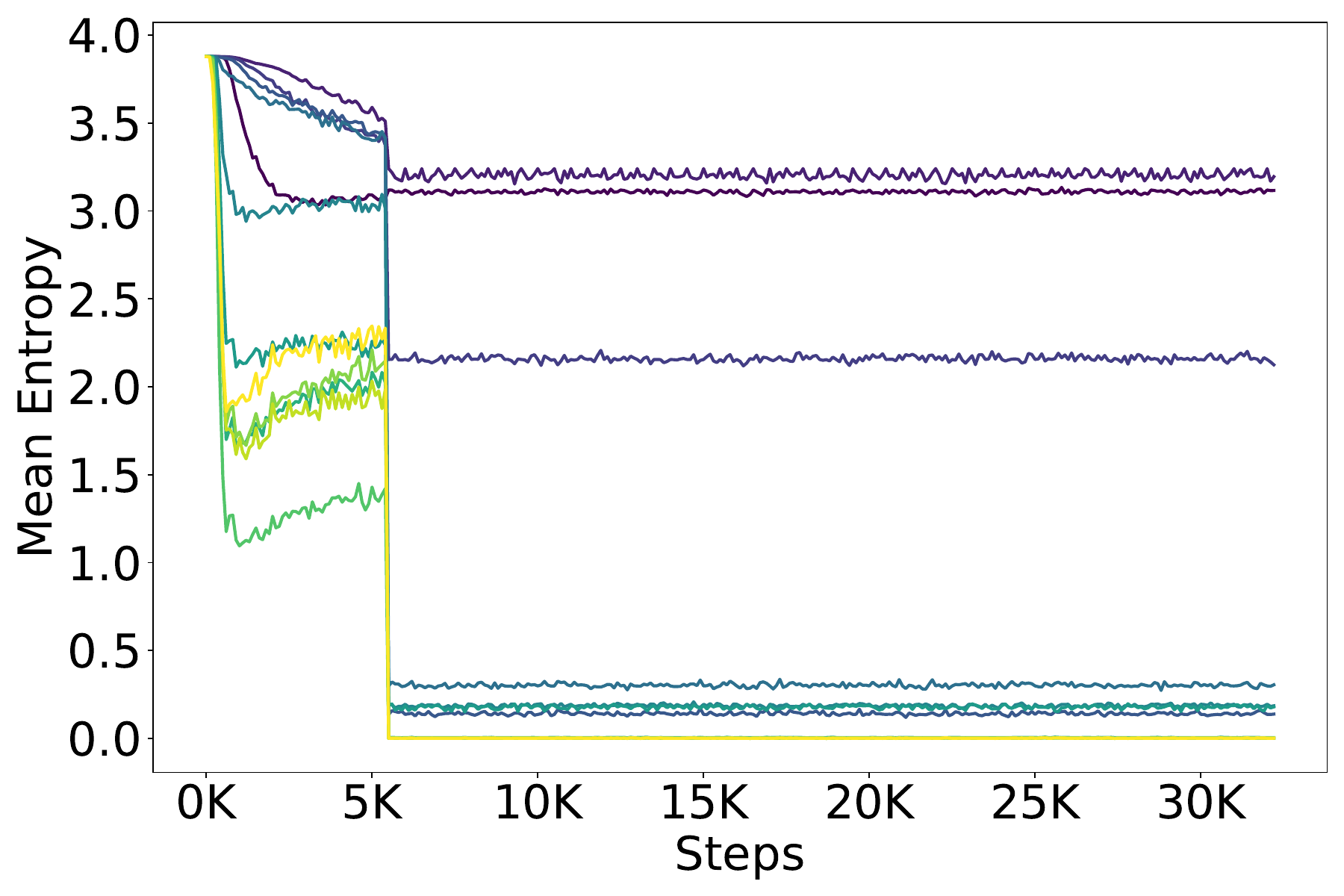}} \\
\vspace{-0.8em}
\subfloat[NaNs observed with a negative slope of $2e$-1 \label{subfig:NaNs2e1}]{\includegraphics[width=.5\textwidth]{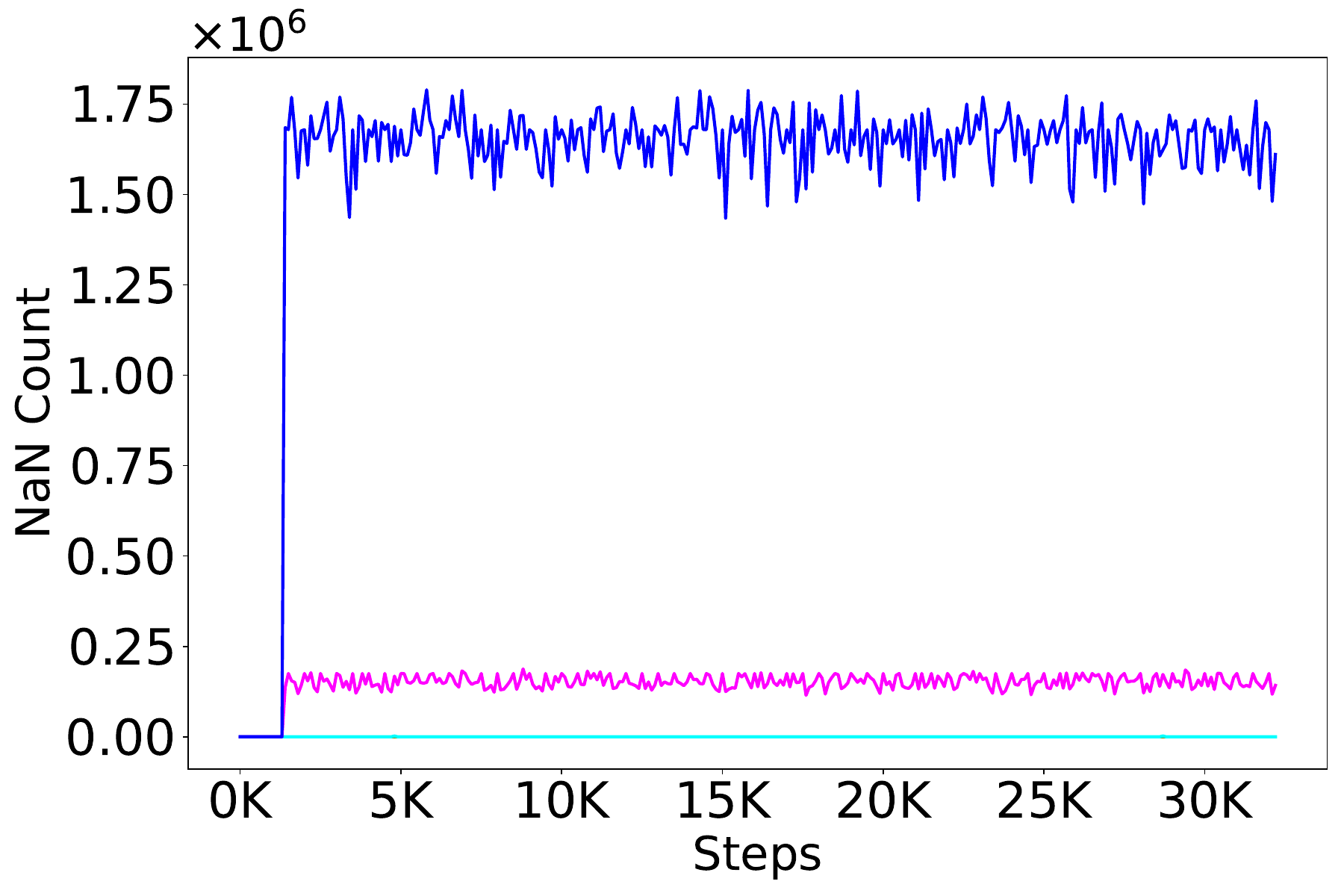}} 
\subfloat[Entropy dynamics with a negative slope of $2e$-1 \label{subfig:Ent2e1}]{\includegraphics[width=.48\textwidth]{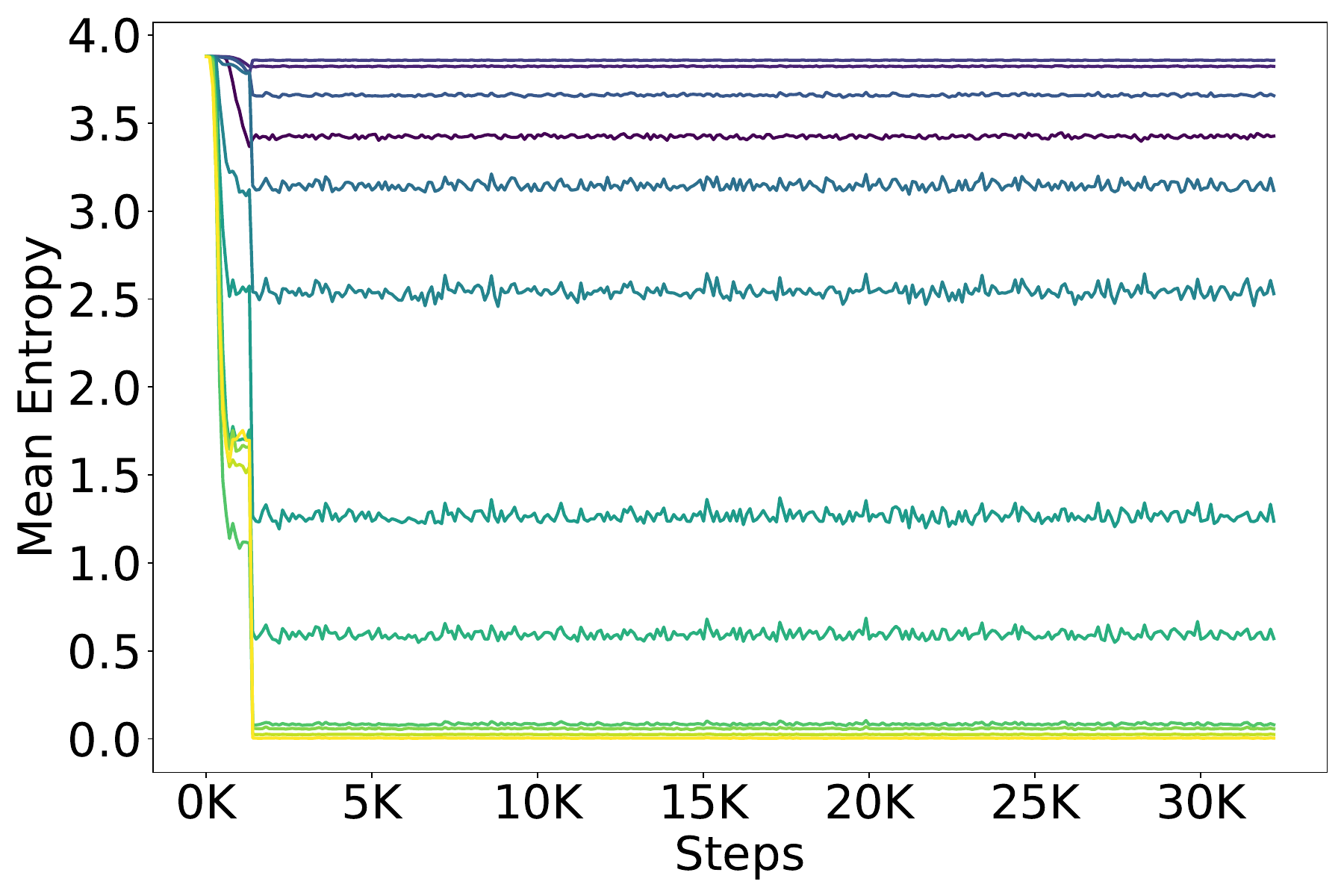}} 
\vspace{-0.2em}
\caption{ Training instability, indicated by  NaNs,  and corresponding entropy dynamics in Normalization-Free GPT-2 ($L$=12, $H$=12, $d$=768) models with fixed negative slopes in the leaky ReLU. The larger the negative slope, the earlier the training instability and entropy collapse occurred. } 
\label{fig:TrainingInstabilityNegSlope}
\end{figure}

%Evolution of Layerwise entropy when GPT-2 ($L$=12, $H$=12, $d$=768) models with various nonlinearity configurations are trained from scratch on CodeParrot dataset. Evolution of layer-wise entropy during training of GPT-2 models ($L$=12, $H$=12, $d$=768) with different nonlinearity configurations on the CodeParrot dataset. The near-identical entropy dynamics in Figures d, e, and f underscore a {\em natural preference} for a zero negative slope, similar to ReLU, in the FFN activation function of the normalization-free model.

{\bf Training instability and entropy dynamics with a fixed negative slope}
In normalization-free LLMs, ReLU-like activation functions, with a near-zero negative slope, naturally stand out as a preferred choice compared to the conventional GELU, offering both improved predictive performance and stable training dynamics. This makes exploring fixed negative slopes in leaky ReLU activations particularly intriguing.

To systematically investigate this, we conducted a series of experiments on normalization-free GPT-2 models, adjusting the negative slopes to fixed values of $1e$-2, $5e$-2, $1e$-1, and $2e$-1. We assessed training instability by monitoring the frequency and distribution of NaN values across model layers and evaluated entropy dynamics across the model’s depth. The results are shown in Figure \ref{fig:TrainingInstabilityNegSlope}. 

For a negative slope of $1e$-2, we observed sporadic occurrences of NaNs primarily in the last layer (Layer 11), as shown in Figure \ref{subfig:NaNs1e2}, with no entropy collapse (Figure \ref{subfig:Ent1e2}). However, as the negative slope increased to $5e$-2, $1e$-1, and $2e$-1, NaNs began to appear consistently in deeper layers, as evidenced by the NaN counts in Figures \ref{subfig:NaNs5e2}, \ref{subfig:NaNs1e1}, and \ref{subfig:NaNs2e1}, respectively. These consistent NaNs correlated with entropy collapses in the deeper layers, indicating a strong relationship between increased negative slope and training instability (Figures \ref{subfig:Ent5e2}, \ref{subfig:Ent1e1}, and \ref{subfig:Ent2e1}).

An interesting trend emerged: the larger the negative slope, the earlier the training instability and entropy collapse occurred. For example, with a slope of $2e$-1, instability occurred almost immediately (Figure \ref{subfig:NaNs2e1}), and entropy collapses in deeper layers (Figure \ref{subfig:Ent2e1}) occurred much sooner compared to lower negative slopes. This suggests that as the slope increases, the window of stable training narrows, making it crucial to choose the appropriate negative slope to avoid early instability and entropy collapse in normalization-free models.

{\bf Broader implications of activation function characteristics in normalization-free models}
The emergence of training instability and entropy collapse at larger negative slopes underscores the sensitivity of normalization-free LLMs to the choice of activation function parameters. The strong correlation between larger negative slopes and earlier training instability suggests that even seemingly minor changes to the negative slope of the leaky ReLU function can significantly influence the model’s ability to maintain stability during training. Specifically, larger negative slopes in leaky ReLU activations {\em aggravate the proliferation of NaNs and exacerbate entropy collapse in deeper layers}, leading to earlier and more pronounced training instability.

This suggests that a near-zero negative slope strikes a crucial balance in normalization-free LLMs, offering sufficient nonlinearity while maintaining training stability. 
\section{Conclusion}
In this paper, we investigated the design of normalization-free decoder-only language models and highlighted the critical role of activation functions in such architectures. Our empirical studies revealed that, contrary to conventional practices, the ReLU activation significantly outperforms, an 8.2\% improvement in perplexity, the GELU in normalization-free models. We found that models with learnable negative slopes in leaky ReLU activations naturally converge toward zero negative slopes, effectively resembling ReLU. 

Additionally, we discovered that LayerNorm-free models with GELU activation suffer from entropic overload in early layers, leading to under-utilization of their representational capacity. These findings underscore the necessity of rethinking activation function choices when LayerNorm is absent and suggest that selecting appropriate activations like ReLU enables the development of transformer models that are more efficient, interpretable, and better suited for applications such as private inference and quantization.

{\bf Limitations.}
This study mainly focuses on pre-training performance, with perplexity as the primary metric, and does not include experiments to evaluate other capabilities such as transfer learning or few-shot learning. Additionally,  our findings are been validated on models with fewer than 1B parameters. Future work will explore broader experimental evaluations,  including the large-scale models (see Appendix \ref{Appendix:FutureWork}).

{\bf Notes.} This workshop submission delves into one of the key findings---the LayerNorm-free design--from our comprehensive paper \href{https://arxiv.org/abs/2410.13060}{AERO: Softmax-Only LLMs for Efficient Private Inference}. The code and implementation are available at \href{https://github.com/Nandan91/relu-revival-normfree}{relu-revival-normfree}.

\bibliographystyle{unsrt}
\bibliography{MyRef}

\newpage
\appendix

\addcontentsline{toc}{section}{Appendix} % Add the appendix text to the document TOC
\part{Appendix} % Start the appendix part
\parttoc % Insert the appendix TOC

\newpage

\section{Why Training from Scratch to Study Nonlinearities?}

Understanding the intricate roles of architectural components and nonlinearities---such as activation functions (e.g., GELU, ReLU) in FFN, normalization layers (e.g., LayerNorm), etc.---in transformer-based language models necessitates a methodical and detailed investigative approach. Training models from scratch is essential for this purpose, as it allows us to delve into the internal mechanisms of the model using quantitative measures like entropy. Below, we present a justification for our methodology:

\begin{itemize} [noitemsep,nolistsep,leftmargin=0.5cm]
\item {\em Nonlinearities' impact on the fundamental learning dynamics:}
Nonlinearities significantly influence the optimization landscape by affecting gradient flow and the model's ability to navigate non-convex loss surfaces. Training models from scratch allow us to observe the fundamental learning dynamics that emerge during the initial stages of training. Thus, constructing models with controlled variations, such as substituting or excluding specific nonlinearities, enables us to isolate their direct effects impact on convergence behavior and training stability.

\item {\em Understanding internal mechanisms through entropy analysis:}
Training from scratch enables us to navigate the evolution of entropy values across the layers and assess how architectural components influence information flow within the model. This analysis provides deep insights into the internal workings of models that may not be accessible when starting from pre-trained checkpoints.

\item {\em Limitations of fine-tuning approaches:}
The aforementioned granular level of analysis is unattainable when starting from pre-trained models, where the optimization trajectory has already been largely determined. In contrast, training models from scratch eliminates confounding variables that could arise from pre-existing weights and learned representations, ensuring that any observed effects are solely due to the architectural modifications introduced.

\end{itemize}

\section{Why Use Entropy to Evaluate the Impact of Nonlinearities?} \label{Appendix:EntropyJustification}
We use entropy as a metric to study the impact of nonlinearities on the transformer-based LLMs for the following reasons: 

\begin{itemize}[noitemsep,nolistsep,leftmargin=0.3cm]
\item {\em Quantifying attention distribution:} As the attention mechanism is fundamental to all transformer-based architecture, computing the entropy of attention score distributions reveals how nonlinearities affect attention concentration. High entropy quantifies exploration and low entropy indicates exploitation. 
\item {\em Feature selection:} Nonlinearities like ReLU enable feature selectivity by amplifying important features and suppressing less relevant ones \citep{maas2013rectifier}. Entropy can measure this selectivity across layers and heads, providing insights into the model's prioritization of information.
Previously, entropy has been used to quantify the layerwise information flow in neural networks \citep{peer2022improving}.
\item {\em Exploration vs. exploitation}: Nonlinear operators like the self-attention mechanism, LayerNorm, and GELU balance exploration and exploitation by selecting relevant features while considering a broader context. For instance, heads in the first layer focus on exploration, while those in the second layer focus on exploitation. (see Figures \ref{subfig:BaselineGELU}, \ref{subfig:BaselineReLU}, \ref{subfig:sm_ln_g} and \ref{subfig:sm_ln_r}). 
\item {\em Systematic assessment:} Prior work \cite{zhang2024the,nahshan2024linear,zhai2023stabilizing,vig2019analyzing,ghader2017does} also used entropy to analyze the behavior of transformer-based models; thus,  enhancing validity and comparability of our findings. 
\end{itemize}

\section{Perplexity as a Reliable Metric to Evaluate the LLMs' Performance}
Perplexity \citep{jelinek1977perplexity} is a widely adopted metric to evaluate the predictive performance of auto-regressive language models, reflecting the model’s ability to predict the next token in a sequence. However, for perplexity to serve as a meaningful comparative metric across different architectures, it is critical to ensure consistency in the tokenizer, and vocabulary size and quality \citep{hutchins2022block}. Any variation in these components can potentially skew the results by inflating or deflating perplexity scores; thus, obfuscating the true effects of architectural changes.

In our work, we maintain tokenization schemes and vocabulary attributes as invariant factors across all experiments within a dataset. This isolation of architectural modifications ensures that any observed variations in perplexity are directly attributable to changes in the model design. Thus, by enforcing a consistent tokenization scheme and vocabulary within a dataset, we ensure that perplexity remains a reliable metric for comparing model architectures. Consequently, lower perplexity in our evaluations reliably reflects improved token-level predictions.

\section{Future Work} \label{Appendix:FutureWork}

{\bf Scaling up and generalizing to larger models}
This research opens several avenues for optimizing LayerNorm-free transformer architectures. A primary direction is scaling up experiments to larger models. Evaluating whether the benefits of ReLU activation persist in models with significantly more parameters will determine the applicability of our findings to state-of-the-art language models. Additionally, extending the analysis to other architectures, such as encoder-only or encoder-decoder transformers, could help generalize our insights across different model types.

{\bf Downstream task performance}
While our study focused on perplexity and entropy metrics, future work should analyze how the choice of activation function affects performance on various downstream tasks. Investigating the implications for fine-tuning processes could also provide valuable insights for practical applications.

{\bf Hybrid activation strategies}
Exploring hybrid activation strategies presents another promising research direction. By using different activation functions in different parts of the model---such as combining GELU in the earlier layers with ReLU in the later layers---we could strike a balance between the benefits observed in our study and the traditional advantages of GELU. This approach may enhance model performance while maintaining computational efficiency.

{\bf Interpretability and practical applications}
Given the observed differences in entropy distribution between ReLU and GELU models, future research could explore how different activation functions impact the interpretability of LayerNorm-free models. This could potentially lead to more explainable model design, addressing a critical need in the field. Integrating these findings into practical applications like private inference and quantization is also promising, as it could improve both model efficiency and security.

{\bf Knowledge distillation for performance enhancement}
By distilling knowledge from a larger, LayerNorm-equipped teacher model to a smaller, LayerNorm-free student model with appropriate activation functions, the performance and generalization capabilities of the student model can be improved. This approach could mitigate any performance gaps arising from the absence of LayerNorm while maintaining the benefits of simplified computation and improved interpretability.

\end{document}